\documentclass[11pt,a4paper]{article}

\usepackage[T1]{fontenc}
\usepackage[utf8]{inputenc}

\usepackage[a4paper,textwidth=15.5cm,top=2.5cm,bottom=2.8cm]{geometry}
\usepackage{microtype}

\usepackage{amsmath}
\usepackage{amssymb}
\usepackage{libertinus}

\usepackage{tabularx}
\usepackage{graphicx}
\usepackage{booktabs}
\usepackage{rotating}
\usepackage{tikz}
\usetikzlibrary{bayesnet, arrows.meta, automata, positioning, calc}
\usepackage[most]{tcolorbox}

\usepackage[numbers,square,sort&compress]{natbib}
\let\cite\citep

\usepackage{url}
\usepackage[colorlinks=true,linkcolor=blue!50!black,citecolor=green!35!black,urlcolor=blue!60!black]{hyperref}
\usepackage[capitalise,nameinlink]{cleveref}

\newtcolorbox{skipbox}[1][]{%
	enhanced, breakable,
	colback=black!4, colframe=black!4, boxrule=0pt, arc=2pt,
	left=8pt, right=8pt, top=6pt, bottom=6pt,
	before upper={\setlength{\parindent}{1.5em}},
	fonttitle=\itshape\small, coltitle=black!70,
	title={You can safely skip this on first reading.}, attach title to upper={\par\smallskip},
	#1}

\definecolor{aicRed}{HTML}{e78284}
\definecolor{aicGreen}{HTML}{a6d189}
\definecolor{aicBlue}{HTML}{8caaee}
\definecolor{aicOrange}{HTML}{e5c890}
\definecolor{aicGray}{HTML}{dce0e8}

\newcommand{\aicSquare}[1]{%
	\tikz[baseline=-3pt]{%
		\node[fill=#1, rounded corners=3pt, minimum size=12pt, draw=none] {};%
	}%
}
\newcommand{\aicCircle}[1]{%
	\tikz[baseline=-3pt]{%
		\draw[fill=#1, draw=none] (0,0) circle (6pt);%
	}%
}
\newcommand{\aicDiamond}[1]{%
	\tikz[baseline=-2.5pt]{%
		\node[
		diamond,
		fill=#1,
		draw=none,
		minimum width=13pt,
		minimum height=13pt,
		inner sep=0pt
		] {};%
	}%
}

\newcommand{\visState}{\aicCircle{aicBlue}}
\newcommand{\visInterconnect}{\aicSquare{aicGreen}}
\newcommand{\visSensor}{\aicCircle{aicOrange}}
\newcommand{\visGoal}{\aicDiamond{aicRed}}
\newcommand{\visGradientPath}{%
	\tikz[baseline={(current bounding box.center)}]{%
		\coordinate (b) at (-30:0.35);
		\draw[-{Latex[scale=0.9]}, color=aicRed, line width=0.75pt]
		(0,0) -- (b);
		\draw[-{Latex[scale=0.9]}, color=aicRed, line width=0.75pt]
		(b) -- ++(10:0.35);
	}%
}

\tikzset{
	aicon/state/.style={
		circle,
		fill=aicBlue,
		minimum size=18pt,
		inner sep=0pt,
		text=white,
		font=\bfseries\footnotesize
	},
	aicon/active/.style={
		rectangle,
		rounded corners=4pt,
		fill=aicGreen,
		minimum size=18pt,
		inner sep=2pt,
		text=white,
		font=\bfseries\footnotesize
	},
	aicon/sensor/.style={
		circle,
		rounded corners=4pt,
		fill=aicOrange,
		minimum size=18pt,
		inner sep=0pt,
		text=white,
		font=\bfseries\footnotesize
	},
	aicon/goal/.style={
		diamond,
		fill=aicRed,
		minimum size=20pt,
		inner sep=0pt,
		text=white,
		font=\bfseries\footnotesize
	},
	aicon/link/.style={
		draw=aicGray,
		line width=2.5pt
	},
	aicon/short/.style={
		line width=1.4pt,
		-{Latex[length=4pt,width=5pt]}
	},
	aicon/gradient/.style={
		draw=aicRed,
		line width=2.5pt,
		->,
		>=latex
	}
}

\newcommand{\drawMyAiconGraph}[2]{%
	\node[aicon/sensor] (#1S1) at (0,4) {$s_1$};
	\node[aicon/sensor] (#1S2) at (0,0) {$s_2$};
	\node[aicon/sensor] (#1A1) at (5,0) {$a$};

	\node[aicon/active] (#1C1) at (0.0, 2) {$c_1$};
	\node[aicon/active] (#1C2) at (1.5, 1) {$c_2$};
	\node[aicon/active] (#1C3) at (5.0, 2) {$c_3$};
	\node[aicon/active] (#1C4) at (3.5, 3.5) {$c_4$};

	\node[aicon/state] (#1x1) at (1.5, 3.5) {$\mathbf{x}_1$};
	\node[aicon/state] (#1x2) at (2.5, 5) {$\mathbf{x}_2$};
	\node[aicon/state] (#1x3) at (5.0, 4.5) {$\mathbf{x}_3$};
	\node[aicon/state] (#1x4) at (3.0, 2.0) {$\mathbf{x}_4$};

	\node[aicon/goal] (#1G) at (5.0, 6.0) {$g$};

	\draw[aicon/link] (#1S1) -- (#1C1);
	\draw[aicon/link] (#1S2) -- (#1C2);
	\draw[aicon/link] (#1A1) -- (#1C3);

	\draw[aicon/link] (#1C1) -- (#1x1);
	\draw[aicon/link] (#1C2) -- (#1x1);
	\draw[aicon/link] (#1C2) -- (#1x4);
	\draw[aicon/link] (#1C3) -- (#1x4);
	\draw[aicon/link] (#1C4) -- (#1x1);
	\draw[aicon/link] (#1C4) -- (#1x2);
	\draw[aicon/link] (#1C4) -- (#1x3);
	\draw[aicon/link] (#1C3) -- (#1x3);
	\draw[aicon/link] (#1x3) -- (#1G);

	\ifnum#2=1
	\draw[aicon/gradient] (#1G) -- (#1x3);
	\draw[aicon/gradient] (#1x3) -- (#1C4);
	\draw[aicon/gradient] (#1C4) -- (#1x1);
	\draw[aicon/gradient] (#1x1) -- (#1C2);
	\draw[aicon/gradient] (#1C2) -- (#1x4);
	\draw[aicon/gradient] (#1x4) -- (#1C3);
	\draw[aicon/gradient] (#1C3) -- (#1A1);
	\fi
}

\newcommand{\drawCubeDrawerNet}[1]{%
	\node[aicon/sensor] (#1Senc)   at (4.5,0) {$s_{\mathrm{enc}}$};
	\node[aicon/sensor] (#1Sforce) at (2.5,0) {$s_{\mathrm{force}}$};
	\node[aicon/sensor] (#1A)      at (6.5,0) {$a$};
	\node[aicon/active] (#1Crob)   at (5.5,1.5) {$c_{\mathrm{rob}}$};
	\node[aicon/active] (#1Cgrasp) at (2.5,1.5) {$c_{\mathrm{grasp}}$};
	\node[aicon/state] (#1Xcube)   at (2.5,5) {$\mathbf{x}_{\mathrm{cube}}$};
	\node[aicon/state] (#1Xee)     at (4.5,3) {$\mathbf{x}_{\mathrm{ee}}$};
	\node[aicon/state] (#1Xdrawer) at (6.5,5) {$\mathbf{x}_{\mathrm{drawer}}$};
	\node[aicon/state] (#1Xpris)   at (4.5,5) {$\mathbf{x}_{\mathrm{pris}}$};
	\node[aicon/state] (#1Pcg)  at (2.5,2.5) {$p_{\mathrm{cg}}$};
	\node[aicon/state] (#1Pdg)  at (6.5,2.5) {$p_{\mathrm{dg}}$};
	\node[aicon/state] (#1Pcid) at (4.5,7) {$p_{\mathrm{cid}}$};
	\node[aicon/active] (#1Ccg)   at (2.5,3.5) {$c_{\mathrm{cg}}$};
	\node[aicon/active] (#1Cdg)   at (6.5,3.5) {$c_{\mathrm{dg}}$};
	\node[aicon/active] (#1Ckin)  at (4.5,4) {$c_{\mathrm{kin}}$};
	\node[aicon/active] (#1Ccid)  at (4.5,6) {$c_{\mathrm{cid}}$};
	\node[aicon/goal] (#1G) at (4.5,8) {$g$};
	\draw[aicon/link] (#1Senc) -- (#1Crob);
	\draw[aicon/link] (#1Sforce) -- (#1Cgrasp);
	\draw[aicon/link] (#1A) -- (#1Crob);
	\draw[aicon/link] (#1Crob) -- (#1Xee);
	\draw[aicon/link] (#1Cgrasp) -- (#1Pcg);
	\draw[aicon/link] (#1Cgrasp) -- (#1Pdg);
	\draw[aicon/link] (#1Ckin) -- (#1Xdrawer);
	\draw[aicon/link] (#1Ckin) -- (#1Xpris);
	\draw[aicon/link] (#1Ccg) -- (#1Xcube);
	\draw[aicon/link] (#1Ccg) -- (#1Xee);
	\draw[aicon/link] (#1Ccg) -- (#1Pcg);
	\draw[aicon/link] (#1Cdg) -- (#1Xdrawer);
	\draw[aicon/link] (#1Cdg) -- (#1Xee);
	\draw[aicon/link] (#1Cdg) -- (#1Pdg);
	\draw[aicon/link] (#1Ccid) -- (#1Xcube);
	\draw[aicon/link] (#1Ccid) -- (#1Xdrawer);
	\draw[aicon/link] (#1Ccid) -- (#1Xpris);
	\draw[aicon/link] (#1Ccid) -- (#1Pcid);
	\draw[aicon/link] (#1Pcid) -- (#1G);
}

\makeatletter
\def\bstctlcite{\@ifnextchar[{\@bstctlcite}{\@bstctlcite[@auxout]}}
\def\@bstctlcite[#1]#2{\@bsphack
	\@for\@citeb:=#2\do{%
		\edef\@citeb{\expandafter\@firstofone\@citeb}%
		\if@filesw\immediate\write\csname #1\endcsname{\string\citation{\@citeb}}\fi}%
	\@esphack}
\makeatother

\DeclareMathOperator*{\argmax}{argmax}

\title{\textbf{Behavior Generation for the Real World:\\ A Tutorial on AICON}}
\author{Vito Mengers\textsuperscript{1,2} \and Adrian Pfisterer\textsuperscript{1} \and Patrick Lowin\textsuperscript{1,3} \and Pu Xu\textsuperscript{1,2} \and Oliver Brock\textsuperscript{1,2,3}\\[6pt]
	\normalsize \textsuperscript{1}Robotics and Biology Laboratory, Technische Universit\"at Berlin, Germany\\
	\normalsize \textsuperscript{2}Science of Intelligence, Research Cluster of Excellence, Berlin, Germany\\
	\normalsize \textsuperscript{3}Robotics Institute Germany}
\date{}

\begin{document}
	\bstctlcite{IEEEexample:BSTcontrol}

	\maketitle

	\begin{abstract}
		AICON (Active InterCONnect) generates behavior by gradient descent through a graph of recursive estimators coupled by differentiable constraints, without searching over future trajectories or hand-specifying controllers and transitions. Across a series of papers, it has produced robust real-world robot systems, including sequential manipulation with emergent interactive perception and failure recovery, multi-goal pushing, and articulated-object perception, and it has served as a model of human vision and behavior. This paper is the first unified account of AICON, written as a tutorial for roboticists and other scientists. It also adds two technical results that go beyond the earlier papers. First, we give one formulation in which a single constraint serves both perception and action: for perception, its residual updates the connected beliefs; for action, differentiating that update yields a factor in the goal gradient. Second, we show that this formulation requires no designed gating or switching: we only write down regularities, and the factor a constraint contributes to the gradient scales with the belief that the regularity currently holds. As beliefs change, some gradient paths steepen and others flatten, and selecting the steepest path yields the emergent subgoal sequencing observed in prior work, which we work through on a cube-in-drawer example. The tutorial part organizes inference schemes for Gaussian, categorical, and particle beliefs into one overview, gives a step-by-step recipe for building an AICON graph for a new problem, outlines interfaces with reinforcement learning and learning from demonstration, relates AICON to planning, control, model predictive control, factor graphs, and active inference, and states its limitations. An open-source implementation is available at  \href{https://github.com/tu-rbo/aicon}{\nolinkurl{github.com/tu-rbo/aicon}}.
	\end{abstract}

	%======================================================================
	\section{Introduction, Overview, Philosophy, History}\label{sec:intro}
	%======================================================================
	%

	This document is a tutorial on AICON (Active InterCONnect), a method to generate behavior that we have developed and applied over a series of papers, from robust real-world robot systems to models of biological behavior (\Cref{sec:preview}). We wrote it because we believe that AICON offers capabilities that complement existing approaches in ways that are worth understanding. The following sections are written for roboticists and other scientists unfamiliar with AICON, as a gentle introduction to its relatively simple but powerful ideas. Our goal is to enable you to understand AICON to the point that you can apply it to your problems. This document, therefore, interleaves conceptual and practical considerations. Our goal is to expose the core ideas so that you can criticize, disprove, improve, exploit, and appreciate them.

	It turns out that AICON captures something of how agents interact with the world, not just robots. We have explored this in biology, psychology, and vision science with some encouraging results, and dedicate a section to readers from those fields. We suspect the ideas may travel further still: into neuroscience, sociology, and beyond, and we would genuinely welcome those explorations.

	An important part of describing AICON will be to point out its limitations. No model can be universal, no model can describe everything well. So we will take extra care in pointing out these limitations (\Cref{sec:limitations}). As you will see, AICON is a useful model for agents that want to interact with the world. Of course, this is the case in robotics. But the disciplines already mentioned above are also concerned with agents that interact with the world. So AICON might be useful there too. Please note that we speak of ``agents'', not of atoms or molecules. So please expect AICON to be useful only at certain levels of abstractions, above physics or chemistry, for example.

	%----------------------------------------------------------------------
	\subsection{What is new about AICON?}\label{sec:whatsnew}
	%----------------------------------------------------------------------

	Most behavior generation approaches commit to a long sequence of actions inside a fixed model of the world. AICON does not. Instead, it continuously updates its model from sensing and, based on that current state, identifies the subgoal that matters next. This is the key to AICON's robustness: it retains the responsiveness of control while extending it to sequential tasks that are usually assigned to planning.

	To make this novelty precise, it is helpful to distinguish between large worlds and small worlds. Leonard Savage made this distinction in his 1954 book ``The Foundations of Statistics''~\cite{savage1954foundations} and had some interesting ideas about it. He was concerned with decision making in the real world, a problem that shares many similarities with behaving in the real world. He realized that any decision making formalism would have to be based on an incomplete model of the world.

	Savage called the real world the ``large world.'' The large world is inherently not modelable by something we can put into a computer or robot: It is impossible to know everything there is to know. To make any kind of progress, he concluded, we must resort to small world models. These are models that pretend to be complete. The methods operating on these models simply assume that the model is complete. And this makes sense, because we cannot do anything about the things we do not know anything about. But this leaves an important gap between the large world and our small world models.

	We argue that this gap is an important contributor to the lack of robustness and generality in today's robotic systems. And we will argue that AICON narrows this gap to an important extent.

\begin{figure}[t]
		\centering
		\begin{tikzpicture}[
			font=\small,
			quad/.style={minimum width=5.4cm, minimum height=3.2cm, align=center, inner sep=4pt, text width=5cm}
			]
			% Quadrants
			\node[quad, fill=aicGray]      (TL) at (0,0)       {\textbf{Planning}\\ task, motion, TAMP\\[3pt] \textbf{Reinforcement learning}\\ (even when data-driven)};
			\node[quad, fill=aicBlue!45]   (TR) at (5.4,0)     {\textbf{\large AICON}};
			\node[quad, fill=aicOrange!30] (BL) at (0,-3.2)    {\textbf{Open-loop execution}\\ of a single precomputed motion\\[3pt] \footnotesize sparsely populated: pursuing one goal\\ \footnotesize without re-sensing is rarely useful};
			\node[quad, fill=aicGreen!45]  (BR) at (5.4,-3.2)  {\textbf{Reactive control}\\ servoing, potential fields};
			% Grid lines
			\draw[white, line width=2pt] (TL.north east) -- (BR.south west -| TL.north east);
			\draw[white, line width=2pt] (TL.south west) -- (TR.south east);
			% Horizontal axis
			\draw[-{Latex[scale=1.1]}, line width=0.8pt, black!60] ($(BL.south west)+(0,-0.35)$) -- ($(BR.south east)+(0,-0.35)$);
			\node[below=0.45cm of BL.south, text width=5cm, align=center, font=\footnotesize] {\textbf{Small world}\\ commits to the internal model,\\ ignores model errors};
			\node[below=0.45cm of BR.south, text width=5cm, align=center, font=\footnotesize] {\textbf{Large world}\\ re-grounded by continuous sensing,\\ accepts model errors};
			\node[font=\small\itshape] at ($(BL.south east)+(0,-2.0)$) {Does the method address the large-world/small-world gap?};
			% Vertical axis
			\draw[-{Latex[scale=1.1]}, line width=0.8pt, black!60] ($(BL.south west)+(-0.35,0)$) -- ($(TL.north west)+(-0.35,0)$);
			\node[left=0.45cm of BL.west, text width=2.6cm, align=right, font=\footnotesize] {\textbf{Myopic}\\ pursues a single goal};
			\node[left=0.45cm of TL.west, text width=2.6cm, align=right, font=\footnotesize] {\textbf{Sequencing}\\ orders subgoals toward an overarching goal};
			\node[font=\small\itshape, rotate=90] at ($(TL.south west)+(-3.6,0)$) {Goal structure};
		\end{tikzpicture}
		\caption{Behavior generation methods differ along two dimensions: whether they account for the gap between the real (large) world and their internal (small-world) model, and whether they sequence subgoals to achieve a higher-level objective. AICON occupies the top right quadrant, combining the gap-awareness of reactive control with the goal-sequencing of planning. Hybrid methods such as model predictive control and closed-loop replanning are discussed in \Cref{sec:mpc}.}
		\label{fig:intro:aicon_vs_others}
	\end{figure}
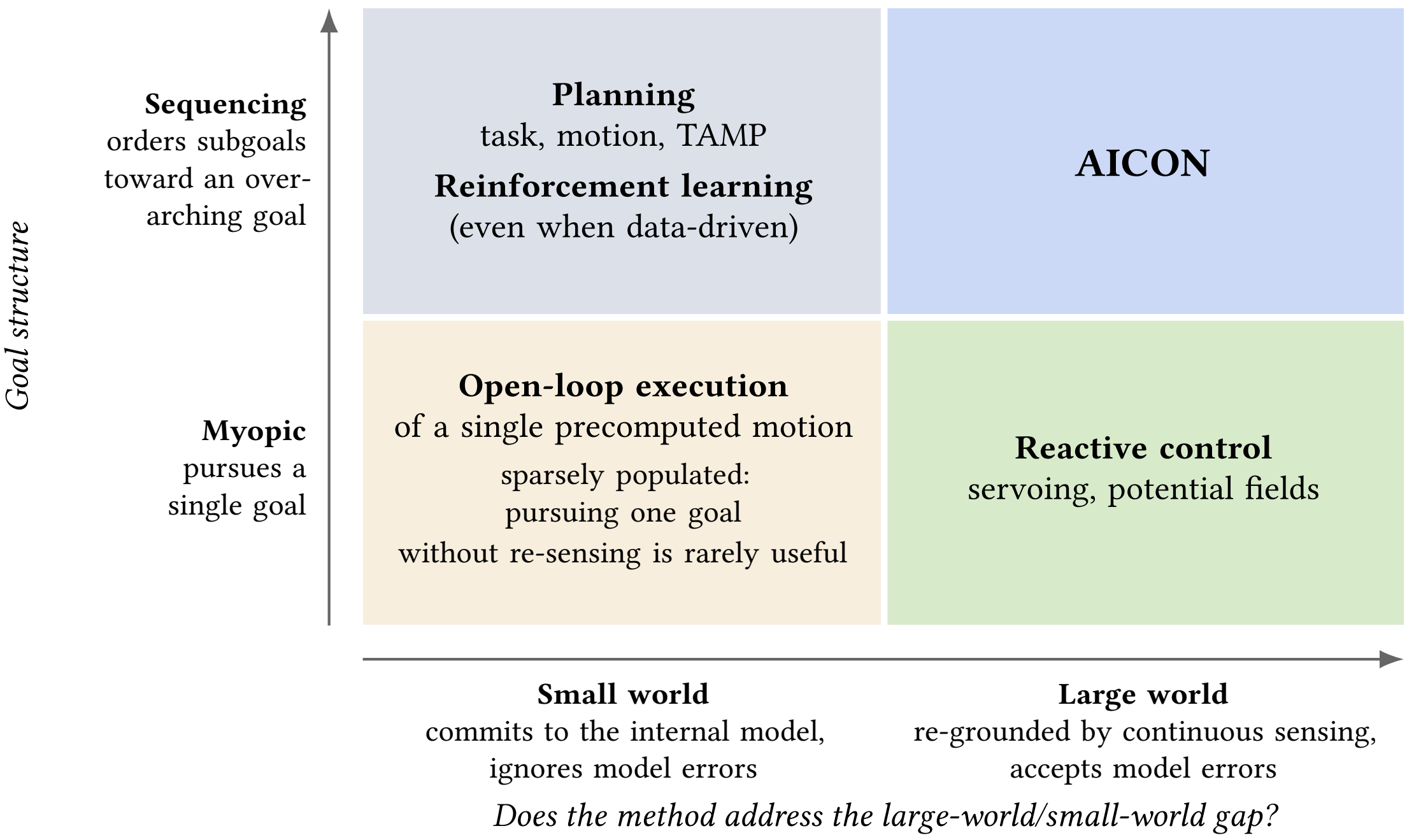

	To pinpoint the novelty of AICON, let us look at \Cref{fig:intro:aicon_vs_others}. We simplify somewhat here to make the point clear, but the broad strokes hold. There are two categorical axes to distinguish between the different behavior generation methods we have in robotics. The horizontal axis distinguishes approaches that do or do not address the gap between the large world and the small world model. The vertical axis distinguishes between approaches that pursue one goal until it is achieved from those that bring intermediate subgoals into the right order to be able to achieve an overarching goal.

	In the bottom right corner, we have reactive control. Reactive control relies on a set of process variables measured directly in the large world. This continuous measurement bridges the gap between the large world and the small world control model.

	In the top left corner, we consider all approaches that explicitly arrange subgoals to achieve the overarching goal, something traditionally associated with planning. In that sense, reinforcement learning (RL) is also planning, as it determines a policy that sequences intermediate subgoals so as to achieve the main goal. We place RL here with some care. It is \emph{not} because RL can never gather enough data about aspects of the world it is unaware of; that argument applies to AICON just as well, since AICON cannot exploit regularities that are not encoded in its graph (\Cref{sec:lim_generalization}). The difference lies in where the ordering of subgoals comes from. An RL policy's ordering is determined at training time, inside the small world of its simulator or dataset, and is then executed as learned. When the deployment world departs from the training distribution, the policy has no mechanism to notice this and re-derive the order of its subgoals. AICON derives this order online from its current beliefs, so any deviation that its estimators capture (a moved object, a failed grasp, a drawer found closed) directly changes which subgoal is pursued.

	The bottom left corner is sparsely populated. It contains methods that pursue a single goal without re-grounding themselves in the large world, for example executing a single precomputed trajectory to a target open-loop. Such methods have the weaknesses of both neighboring quadrants and few of their strengths, which is why they are mostly used where the world is so well controlled that the small world model is effectively complete, as on an assembly line.

	To act well in the real world, both aspects are necessary. An autonomous agent must be aware of the large world rather than just of a small world model. And the agent must be able to handle multiple subgoals in the right order so as to achieve a high-level objective. That means that competent agents should be able to occupy the top right quadrant of \Cref{fig:intro:aicon_vs_others}.

	AICON occupies the top right corner: It is capable of ordering multiple subgoals on the way to achieving a behavioral objective. But it relies on a world model that is continuously updated from sensor measurements of the real world. This promises to combine the advantages of control and planning. In the remainder of this document, we show how it does so and where this combination reaches its limits. AICON bridges the gap between the large world and the small world through continuous sensing. And it enables---in response to the ever-changing world state---identifying and ordering the necessary subgoals to robustly achieve a behavioral objective.

	\paragraph{What this tutorial contributes.}
	AICON has so far been described piecemeal, across papers that each address one problem. This tutorial collects it into a single account. Specifically, it provides:
	\begin{itemize}
		\item a unified formulation in which one differentiable constraint is read out both for perception, as a belief update, and for action, as a factor in the goal gradient (\Cref{eq:readout});
		\item a derivation showing that subgoal sequencing emerges from the encoded regularities alone: beliefs implicitly modulate the steepness of gradient paths, and steepest-path selection does the rest, without any designed gates or transitions, worked through on a cube-in-drawer example (\Cref{sec:actions}, \Cref{fig:sequencing});
		\item an overview of inference schemes for heterogeneous belief representations (\Cref{tab:aicon_table});
		\item a step-by-step guide for building an AICON graph for a new problem (\Cref{sec:howto});
		\item a discussion of how AICON interfaces with learning (\Cref{sec:interfacing}) and relates to other approaches, together with an explicit account of its limitations (\Cref{sec:limitations}).
	\end{itemize}

	%----------------------------------------------------------------------
	\subsection{Preview of what AICON can do}\label{sec:preview}
	%----------------------------------------------------------------------

	Before diving into how AICON works, here is an overview of what it has been used for. Across all of these applications, the key idea is the same: leverage the relationships between quantities in the world to make perception and action more robust. These relationships let AICON resolve ambiguities that no single source of information could resolve, or generate the actions that resolve them. Videos and further material for many of these examples are collected on the AICON project page.\footnote{\url{https://rbo.gitlab-pages.tu-berlin.de/pages/aicon/}}

	\paragraph{Perception.}
	Many perception systems train monolithic models or fuse information from different sources to estimate certain quantities in the world. They rarely leverage the relationships between these quantities to enhance their estimates. AICON has been used for perception tasks that can be decomposed into subproblems with a common structure, which lets estimators exchange information and resolve ambiguities that cannot be disambiguated from a single modality. This has led to robust perception for estimating kinematic structure, including from interaction with articulated objects~\cite{martinmartin2016integrated,martinmartin2022coupled}, from observed human hand motion~\cite{pfisterer2025helping}, and from sound~\cite{baum2023estimating}, as well as for object segmentation~\cite{mengers2023combining} and affordance estimation~\cite{lowin2026coupled}. In all of these examples, complementary information refines the estimates; for instance, the motion of points is used to estimate rigid body poses, while the rigid body poses in turn reduce measurement errors for point motion.

	\paragraph{Behavior generation.}
	Choosing the right action at the right time requires integrating information from different modalities, reasoning about uncertainty, and prioritizing different subgoals. AICON combines all of this in a single framework: by computing the next action via gradient descent through its graph, it shows emergent abilities such as interactive perception, error recovery, and the sequencing of subgoals. We have demonstrated this on sequential tasks such as blocksworld and drawer opening on a real robot~\cite{mengers2025noplan}, on object pushing and other tasks with conflicting goals~\cite{mengers2026riding}, on generalization to new task variants through adaptive compositions of regularities~\cite{battaje2026building}, and, combined with learning, on multi-robot tasks such as object handover, collaborative search, and a pressure-plate task~\cite{sebastian2026world}.

	\paragraph{Modeling biological behavior.}
	AICON-generated behavior also shares similarities with biological behavior. We have used AICON to model human visual illusions~\cite{battaje2024information}, human gaze in dynamic scenes~\cite{mengers2025scanpath}, human planning failures on a clinical task~\cite{migacev2026planhumanreactiverobotics}, and collective behavior in groups of agents~\cite{mengers2024leveraging,mengers2026mechanistic}. We discuss these parallels in \Cref{sec:biology}.

	\paragraph{Software.}
	This document is conceptual. Readers who want an implementation can find an open-source Python framework, with simulated and real-robot examples and a step-by-step tutorial that builds a drawer-opening system, at \url{https://github.com/tu-rbo/aicon}.

	%----------------------------------------------------------------------
	\subsection{Historical context}\label{sec:history}
	%----------------------------------------------------------------------

	\begin{skipbox}
		Historically, the distinction between large worlds and small worlds has not been made explicitly in robotics. Behavior generation was roughly divided into planning and control, and the two were distinguished only along the vertical axis of \Cref{fig:intro:aicon_vs_others}: planning sequences subgoals, control pursues a single goal. Implicitly, planning was located in the small world and control in the large world, so the landscape of behavior generation approaches, at least conceptually, consisted only of the top left and bottom right corners. Adding the large/small world distinction as an independent, horizontal axis splits this landscape into four quadrants and reveals two new ones. AICON lives in one of them. Since AICON is very simple and in retrospect seems rather obvious, the fact that the large/small world distinction was not made explicitly might explain why nobody bothered looking for an approach in the top right square. And neither did we. We basically ``discovered'' AICON almost myopically by the continuous expansion of what we called interactive perception~\cite{katz2008manipulating,martin2014online,martinmartin2022coupled,mengers2025noplan}.

		It is surprising that the distinction between large and small worlds has not played a larger role in robotics until now. This is all the more surprising once you realize that the ``predecessors'' of robotics, the cyberneticists, were acutely aware of this distinction and even pointed out that a model to generate behavior in the real world must contain a model of the relevant aspects of that world. Cyberneticists even proved that \emph{every good regulator of a system must have access to a model of that system}~\cite{Conant-1970,Scholten-2010,Baez-2016,Wentworth-2021}. More recently, \citet{Richens-2025} showed formally that any agent able to generalize across multi-step goal-directed tasks must have learned a predictive model of its environment, and concluded that general agents cannot avoid the problem of world modeling (see also~\cite{Khetarpal-2026}).

		We feel that our distinction between large worlds and small worlds is supported by the work of cyberneticists from the 70s and from deep learning researchers of today. We find ourselves in good company.

		And, of course, we must acknowledge that in robotics there has been a long debate about the role of models. This discussion anticipated in many ways the distinction between large and small worlds. Rod Brooks famously said that the world is its own best model~\cite{brooks1990elephants}, advocating to bridge the large/small-world gap by continuously referring back to the large world. AICON does exactly this.

		In response to Rod Brooks' advocating for using the world as its own model, Matt Mason pointed out that one should not sense that which can more easily be predicted~\cite{mason1993kicking}, advocating for the usefulness of models. AICON does this too. AICON makes extensive use of sensing and model-based prediction.

		Both Rod Brooks and Matt Mason were right, of course. And we hope that AICON is the glue that connects and unifies these two perspectives.

		While on the subject of Rod Brooks: We cannot deny that AICON shares some important properties with Rod Brooks' subsumption architecture~\cite{brooks1986robust}. We fully acknowledge these similarities and are proud to stand on the shoulders of giants. After reading the document, you will hopefully agree that AICON also differs in important respects from the subsumption architecture. But we are happy if you will see AICON as standing in the tradition established in the subsumption architecture several decades ago.
	\end{skipbox}

	%======================================================================
	\section{Introductory Example: Control, Planning, AICON}
	\label{sec:intro_examples}
	%======================================================================

	Agents that want to act in the large world must first generate actions within their small-world model. How agents generate such actions is a central topic in robotics. In this section, we explain how AICON differs from established behavior-generation methods, such as control and planning.
	By working out the differences and similarities, we situate AICON in the landscape of related literature and hopefully enable every reader to compare and contrast to something they know relatively well already.

	Here, planning and control are not presented as a strict classification of methods, but rather as two general paradigms/patterns for behavior generation. Most robot systems in practice employ a variation or combination of both; we discuss hybrids such as model predictive control in \Cref{sec:mpc}.

	Let us consider a simple example. A robot must place a cube inside a drawer. Playing through how different behavior generation paradigms solve this task will allow us to highlight advantages and limitations. We then will use this as a basis to introduce AICON, positioning it relative to these advantages and disadvantages.

	%----------------------------------------------------------------------
	\subsection{Planning}\label{sec:ex_planning}
	%----------------------------------------------------------------------

	Planning, in its most general form, searches for a state-action sequence from a start state of the robot to a desired goal state while satisfying constraints~\cite{ghallabAutomatedPlanningTheory2004,lavallePlanningAlgorithms2006}. Since the notion of state is rather general and can span many different levels of abstractions, different forms of planning have evolved, each at its own abstraction.

	\textit{Motion planning} considers robots as kinematic objects and determines a plan for the motion of this object through a possibly dynamic environment~\cite{lavallePlanningAlgorithms2006}. States describe the physical arrangement of the robot and the objects in the environment. \textit{Task planning}, on the other hand, operates at a higher level of abstraction, and generally describes the state of the world in terms of relationships and first-order logical predicates~\cite{ghallabAutomatedPlanningTheory2004}. A third, more recent approach to planning combines motion and task planning into TAMP (\textit{task and motion planning})~\cite{garrettIntegratedTaskMotion2021}.

	In practice, all of these approaches confront similar challenges that arise from the high dimensionality of the state space and the complex search landscape that encodes preferences over valid solutions.

	Typically, at the task level, the problem is represented symbolically, for example, using STRIPS-style~\cite{fikesSTRIPSNewApproach1971} action schemas:

	\begin{verbatim}
		Action: move_cube(table -> drawer)
		Preconditions: drawer_is_open, holding(cube)
		Effects: at(cube, drawer), not at(cube, table)
	\end{verbatim}

	\noindent A task planner searches over these symbolic actions to produce an action sequence:

	\begin{verbatim}
		grasp(drawer_handle)
		open_drawer
		release(drawer_handle)
		pick(cube, table)
		move_cube(table -> drawer)
		place(cube, drawer)
	\end{verbatim}

	The motion planner then searches for geometrically feasible trajectories to realize each action -- collision-free paths, valid grasps, reachable configurations. If no feasible motion exists for a planned action, the system must backtrack and revise the task plan.

	%----------------------------------------------------------------------
	\subsubsection*{Advantages of Planning}
	%----------------------------------------------------------------------
	\begin{itemize}

		\item \textbf{Global optimality}: Planning over the entire course of action enables searching for globally optimal solutions, rather than greedily optimizing at each step. Because the full trajectory is considered, planning can avoid local minima that trap reactive methods, find time-optimal plans, and resolve long-horizon dependencies.

		\item \textbf{Compositionality}: Planning algorithmically searches the combinatorial space of actions and composes operators that achieve subgoals into a sequence that achieves the goal. The cube-in-drawer task plan illustrates this. Changing the goal---say, moving the cube to a shelf instead---requires only extending the domain description with the relevant predicates and operators, not a redesign of the entire system.

	\end{itemize}

	%----------------------------------------------------------------------
	\subsubsection*{Disadvantages of Planning}
	%----------------------------------------------------------------------
	\begin{itemize}
		\item \textbf{Computational cost}: Planning searches over state-action sequences, and the size of this search space grows exponentially with the dimensionality of the state and action spaces---a problem known as the curse of dimensionality. In our example, for the same cube-in-drawer task, moving from a 6-axis robot arm with a parallel-jaw gripper to a full humanoid with two dexterous hands would increase the planning cost exponentially.
		\item \textbf{Model dependency}: Planning requires a forward model that predicts action outcomes, and plan quality is bounded by model accuracy. Prediction errors also compound over the planning horizon. In the cube-in-drawer example, accurate models involving frictional contact are notoriously difficult to obtain.
		\item \textbf{Brittleness of plan execution}: Plans prescribe a fixed action sequence computed before execution. Any deviation from the expected state---due to perception errors, failures, or environmental changes---can invalidate the rest of the plan. Online replanning and belief-space planning address this by re-perceiving and reasoning over uncertainty, but each extension adds significant computational cost.

	\end{itemize}

	%----------------------------------------------------------------------
	\subsection{Control}\label{sec:ex_control}
	%----------------------------------------------------------------------

	Another common paradigm of behavior generation is control. Instead of planning a full state trajectory that reaches the goal, we use control laws to drive the current state towards the goal states during execution. For complex tasks like our example, instead of using one monolithic control law, we usually partition the state space into discrete regions and transitions, and assign simple control laws, each responsible for one region or transition. This can be implemented via many methods, such as state machines, behavior trees~\cite{colledanchiseBehaviorTreesRobotics2018}, hybrid automata~\cite{henzingerTheoryHybridAutomata2000} and behavior-based robotics~\cite{BehaviorbasedRobotics}. Let's use the simplest example, a finite state machine.

	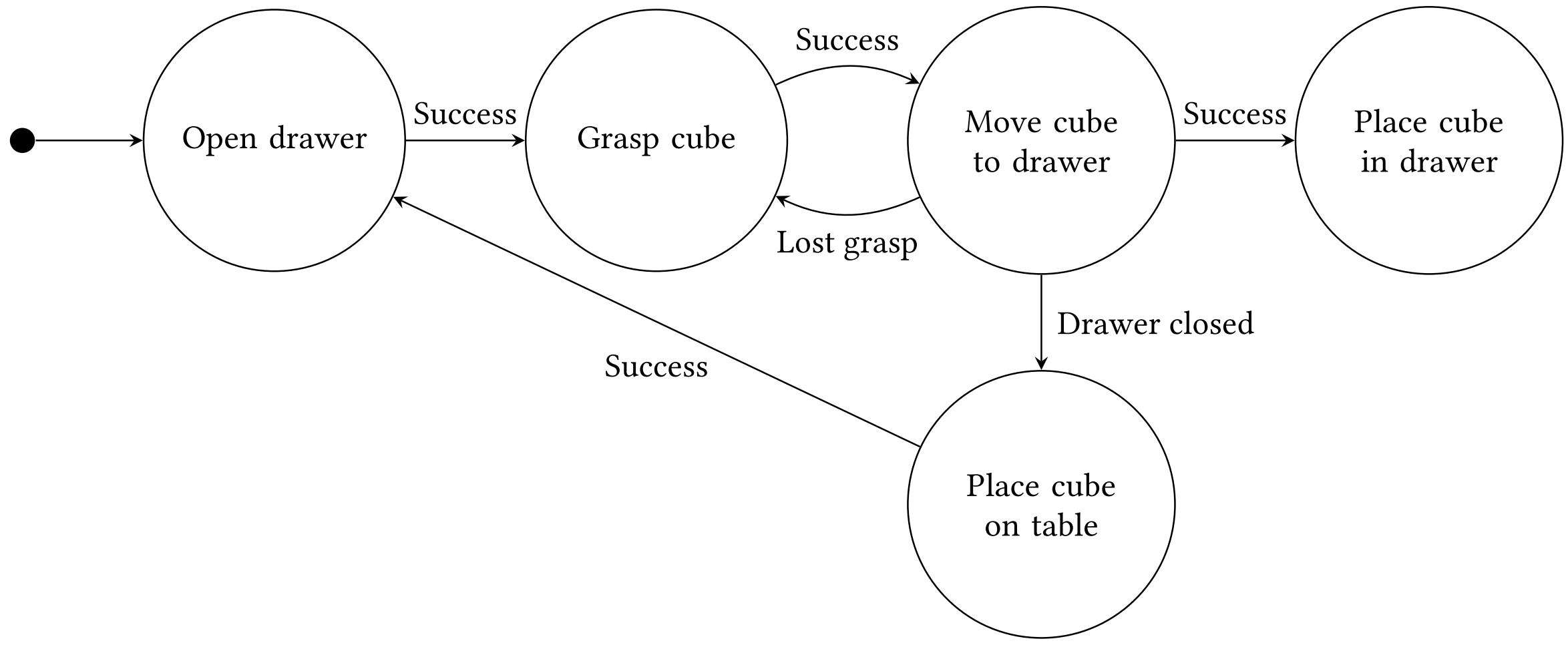
\begin{figure}[t]
		\centering
		\resizebox{\linewidth}{!}{%
			\begin{tikzpicture}[
				->,
				>=stealth,
				font=\scriptsize,
				node distance=0.8cm,
				state/.style={
					draw,
					circle,
					minimum size=2.2cm,
					text width=1.85cm,
					align=center
				},
				start/.style={
					circle,
					fill,
					inner sep=0pt,
					minimum size=6pt,
					align=center
				}
				]

				\node[start] (start) {};
				\node[state, right=0.9cm of start] (open_drawer) {Open drawer};
				\node[state, right=1.0cm of open_drawer] (grasp) {Grasp cube};
				\node[state, right=1.0cm of grasp] (transport) {Move cube to drawer};
				\node[state, right=1.0cm of transport] (place_drawer) {Place cube in drawer};
				\node[state, below=0.8cm of transport] (place_table) {Place cube on table};

				\path

				(start) edge (open_drawer)
				(open_drawer) edge node[above] {Success} (grasp)

				(transport) edge node[above] {Success} (place_drawer)
				(transport) edge node[right] {Drawer closed} (place_table)

				(place_table) edge node[below=4pt] {Success} (open_drawer)

				;

				\path
				(grasp) edge[bend left=25] node[above] {Success} (transport)
				(transport) edge[bend left=25] node[below] {Lost grasp} (grasp)
				;

			\end{tikzpicture}
		}% end resizebox
		\caption{Finite state machine for the cube-in-drawer task. Each node is a discrete world state with an associated control law; edges are transitions triggered by observed conditions. The lower branch handles the failure case in which the drawer is found closed on arrival.}
		\label{fig:state_machine}
	\end{figure}

	\Cref{fig:state_machine} shows a finite state machine for the cube-in-drawer task. Each node represents a discrete region of the world state with a dedicated control law. The robot executes that law until a sensed condition triggers a transition to the next state, thereby generating the overall task behavior from a composition of simple controllers.

	%----------------------------------------------------------------------
	\subsubsection*{Advantages of Control}
	%----------------------------------------------------------------------
	\begin{itemize}
		\item \textbf{Computationally efficient}: Reactive control computes actions using only the current state, avoiding the costly prediction and search over future states that planning requires. This makes control-based methods significantly faster at execution time, enabling real-time operation at high frequencies.
		\item \textbf{Closed-loop robustness}: closed-loop control laws can react to unmodeled disturbances using sensory feedback. For example, as the robot approaches the cube through visual servoing control, the robot can still track the object even if a person moved the cube during execution. A pre-planned trajectory would fail due to this disturbance.
	\end{itemize}

	%----------------------------------------------------------------------
	\subsubsection*{Disadvantages of Control}
	%----------------------------------------------------------------------
	\begin{itemize}
		\item \textbf{Prone to local minima}: Because control laws act on the current state, they are susceptible to local minima. This is a well-known problem in potential field methods~\cite{khatibRealtimeObstacleAvoidance1986}. In the cube-in-drawer example, the robot might get stuck navigating around complex obstacles, endlessly oscillating or halting.
		\item \textbf{Less expressive}: Expressing complex behaviors using control alone is difficult. For structured approaches like finite state machines and behavior trees, the designer must fully specify all states, transitions, failure conditions, and the individual controllers for each state. This manual enumeration is tedious, error-prone, and scales poorly with task complexity. More importantly, the resulting system can only handle situations that were explicitly anticipated. An unspecified state or a missing transition can lead to failure.
	\end{itemize}

	%----------------------------------------------------------------------
	\subsection{AICON Paradigm}\label{sec:ex_aicon}
	%----------------------------------------------------------------------

	After introducing the general philosophy of AICON, we now explain in more detail how it generates behavior. AICON does so without searching over future trajectories and without manually specifying controllers or transitions. Instead, it encodes task-relevant regularities in a network of recursive estimators and active interconnections, and derives actions from this graph via gradient descent.

	Before introducing goals and actions, AICON is a state estimation system. It consists of a network of recursive estimators and interconnections. Each estimator runs its own loop, continuously estimating a state quantity from sensor inputs and information received from connected estimators. The interconnections are differentiable constraint functions between estimated quantities, through which estimators exchange information. For example, in the cube-in-drawer task, one estimator tracks the hand pose and another estimates how far the drawer is open. An interconnection between them encodes the constraint that, when the hand grasps the drawer, they move together. Both estimators use this constraint to exchange information and refine their estimates.

	How does this state estimation system become a behavior generation system? First, we define a goal $g$---a scalar cost function over some estimated state to be minimized. For example, opening the drawer means minimizing the negative opening distance. Then, the action $a$ is simply a quantity in the network that we can control directly---in this case, the end-effector velocity. Achieving the goal then reduces to gradient descent: at each step, move the action along the steepest descent direction of $g$ with respect to $a$, similar to reactive control. What sets AICON apart is how the gradients are calculated and selected. Underlying AICON is a graph structure in which quantities---goals, actions, and states---are sparsely connected by constraints. The edges of this graph are these constraints, and they are read out differently for perception and for action. For perception, a constraint tells the connected estimators how to update their beliefs so that they remain consistent with each other. For action, the same constraint tells us how a change in one quantity would propagate to another, that is, it provides a local derivative between the connected states (we make this precise in \Cref{sec:actions}). A path through this graph from action to goal composes these local derivatives via the chain rule, yielding one possible gradient $\partial g / \partial a$. Multiple paths exist simultaneously, each representing a distinct way the action can influence the goal. The magnitude of the gradient along each path varies with the current state, and AICON follows the steepest one.

	This is how sequencing emerges. Many regularities hold only under certain conditions, and the graph keeps a belief over whether they currently hold. The constraint that couples cube and hand, for example, encodes that a grasped cube moves with the hand. It therefore transmits a derivative only to the degree that the robot believes it holds the cube. Nobody designs this as a switch; it follows from the belief update. In the example of putting a cube in a drawer: when the drawer is open and the cube is on the table, the path in which the hand carries the cube is almost closed, because the grasp belief is low. The steepest path is instead

	\vspace{6pt}

	\noindent\colorbox{gray!15}{action} $\rightarrow$ \colorbox{gray!15}{end-effector pose} $\rightarrow$ \colorbox{gray!15}{$p(\text{cube grasped})$} $\rightarrow$ \colorbox{gray!15}{cube position} $\rightarrow$ \colorbox{gray!15}{$p(\text{cube in drawer})$}

	\vspace{6pt}

	\noindent causing the robot to approach and grasp the cube: raising the grasp belief is currently the most effective way to make the cube movable. Once the cube is grasped, the grasp belief saturates, this path flattens, and the steepest path shifts to

	\vspace{6pt}

	\noindent\colorbox{gray!15}{action} $\rightarrow$ \colorbox{gray!15}{end-effector pose} $\rightarrow$ \colorbox{gray!15}{cube position} $\rightarrow$ \colorbox{gray!15}{$p(\text{cube in drawer})$}

	\vspace{6pt}

	\noindent causing the robot to carry the cube toward the drawer. ``Grasp the cube'' thus appears as a subgoal without ever being specified as one. If the drawer is closed, the same logic first produces ``grasp the handle'' and ``open the drawer''. We work through this in detail in \Cref{sec:actions} and \Cref{fig:sequencing}.

	If this sounds familiar to you, please check out~\Cref{sec:relationship} for a discussion of the differences between AICON and other methods.

	%----------------------------------------------------------------------
	\subsubsection*{Advantages of AICON}
	%----------------------------------------------------------------------

	\begin{itemize}
		\item \textbf{Like control, but more expressive}: AICON shares many similarities with reactive control. But unlike control, where states, transitions, and controllers need to be explicitly enumerated and specified, AICON derives its behavior from a description of the world---estimators that track task-relevant states and constraints that encode relevant regularities. Behaviors emerge from the interaction of these components without being explicitly programmed.
		\item \textbf{No plan, but compositional}: AICON does not search over future state-action trajectories like planning does. But by following the current gradient path that connects the action to the goal, AICON exhibits compositional generalization similar to planning.
	\end{itemize}

	%----------------------------------------------------------------------
	\subsubsection*{Disadvantages of AICON}
	%----------------------------------------------------------------------

	\begin{itemize}
		\item \textbf{No optimality guarantees}: AICON finds \emph{a} solution, not the \emph{best} one. Planning, MDP, and POMDP methods can, under correct models, guarantee time-optimality or minimal cumulative cost. AICON cannot.

		\item \textbf{No completeness}: AICON resolves what is in conflict \emph{now}, but cannot detect that a current action forecloses a necessary future one. Tasks like Tower of Hanoi, which require a deliberate temporary setback, remain out of reach. Planning methods handle these by construction.
	\end{itemize}

	We discuss these and further limitations to AICON in depth in \Cref{sec:limitations}.

	%======================================================================
	\section{The AICON Framework}
	%======================================================================
	\label{sec:aicon}
	This section turns the AICON paradigm into an implementable graph for perception and behavior generation. In AICON, the problem is formulated in terms of two fundamental structural components: \emph{states} and \emph{constraints}. Rather than explicitly specifying behaviors or task-specific logic, we describe how the world evolves. Concretely, we identify (i) the relevant states that characterize the system and (ii) the constraints that must hold between them.

\begin{figure}[t]
		\centering
		\begin{tikzpicture}[scale=0.85, every node/.append style={transform shape}]
			\begin{scope}[shift={(0,0)}]
				\drawMyAiconGraph{L}{0}
				\node[below, font=\bfseries] at (2.5,-0.6) {(a) System topology};
			\end{scope}
			\begin{scope}[shift={(8,0)}]
				\drawMyAiconGraph{R}{1}
				\node[below, font=\bfseries] at (2.5,-0.6) {(b) One possible gradient path};
			\end{scope}
		\end{tikzpicture}
		\caption{A generic AICON graph. (a) The connections between recursive estimators ($\mathbf{x}$), active interconnections ($c$), sensors ($s$), and actuators ($a$). (b) The goal gradient is backpropagated to the actuator $a$ along one possible path.}
		\label{fig:generic_graph}
	\end{figure}
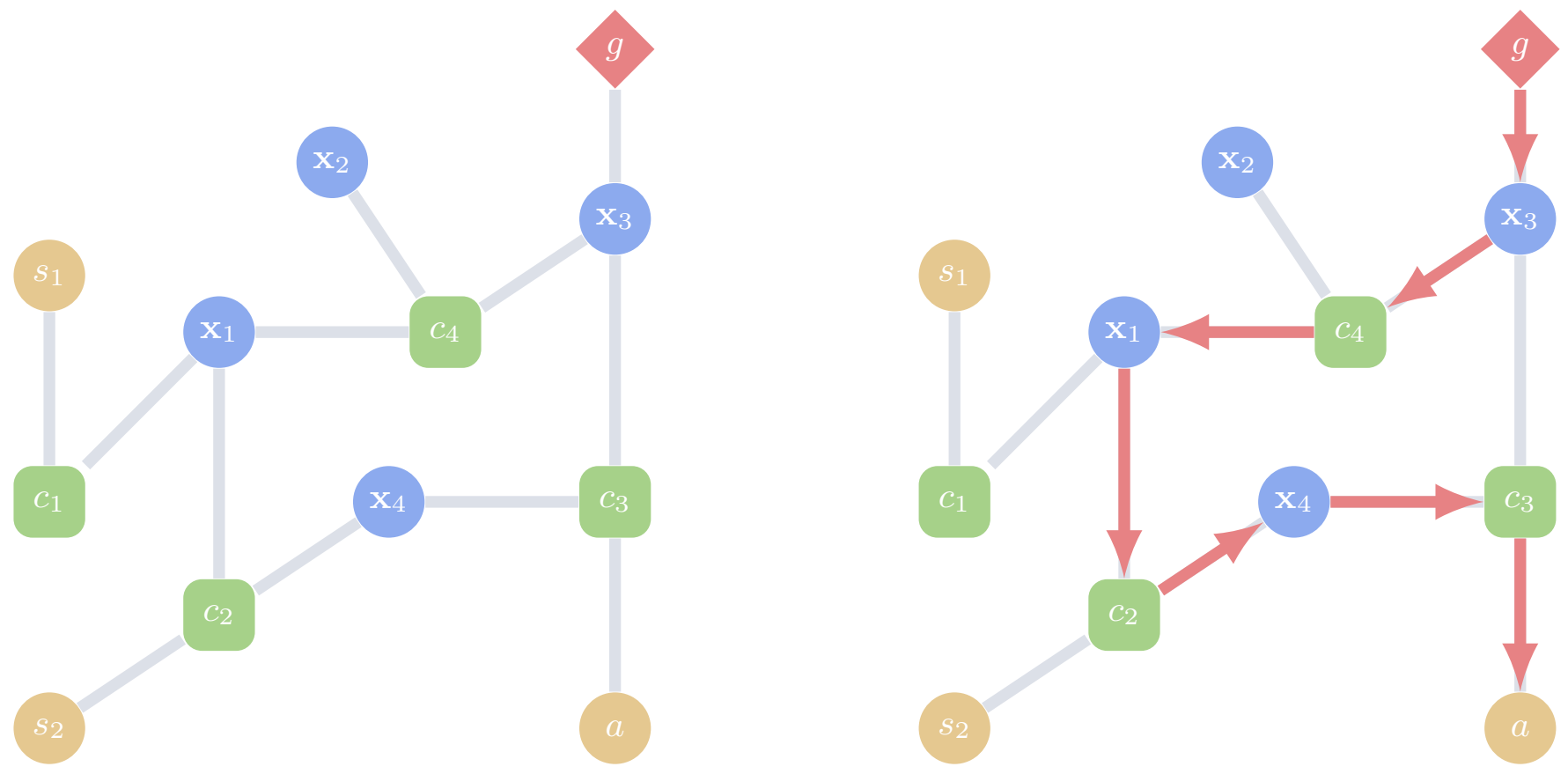

	In the following, we explain how to identify these components and implement them within AICON: states are represented as recursive estimators (\Cref{sec:rec_est}), and constraints are encoded as active interconnections (\Cref{sec:act_int}). Once this structure is in place, both perception and behavior generation follow naturally. Perception is performed via constrained probabilistic inference (\Cref{sec:inference}), while behavior generation is achieved by defining a scalar cost function and performing gradient descent through the system (\Cref{sec:actions}). This explanation will progressively build up the mathematical and graphical notation for AICON which we summarize in \Cref{tab:aicon_components}. \Cref{fig:generic_graph} shows how these components fit together in a generic graph, together with one gradient path.

	\begin{table}[t]
		\centering
		\renewcommand{\arraystretch}{1.8}
		\setlength{\tabcolsep}{8pt}

		% Calculation: 0.7 + 1.3 = 2.0 (The sum must equal the number of X columns)
		\begin{tabularx}{\textwidth}{@{}
				>{\hsize=0.7\hsize}X   % Adjust 0.7 to change Name width
				c
				c
				>{\hsize=1.3\hsize}X   % Adjust 1.3 to change Description width
				@{}}
			\toprule
			\textbf{Component Name} & \textbf{Notation} & \textbf{Symbol} & \textbf{Description} \\
			\midrule

			State / Recursive Estimator & $X, \, \mathbf{x}_t$ & \visState &
			A recursively estimated belief over a world state (e.g., pose, joint angle) updated at time $t$. \\

			Forward Model & $\mathbf{f}(\mathbf{x}_{t-1}; \dots)$ & [incl. above] &
			A differentiable function containing system priors that updates state estimates based on previous states. \\

			Active Interconnection & $\mathbf{c}(\mathbf{x}_1, \dots, \mathbf{x}_N)$ & \visInterconnect &
			A differentiable, implicit function regulating information flow between recursive estimators based on system state and uncertainty. \\

			Sensors / Actuators & $S,\, A, \, \mathbf{a}$ & \visSensor &
			Raw inputs ($S$) and actuation signals ($\mathbf{a}$) that influence the belief graph. \\

			Goal & $g(\mathbf{x})$ & \visGoal &
			A differentiable scalar cost function on a state or belief, defining the objective to minimize. \\

			Gradient Path & $\nabla^{\mathrm{p}}_{\mathbf{a}}\,g$ & \visGradientPath &
			Topological path $p$ of steepest descent. It chains the local derivatives that the constraints along the path induce between states, backpropagating the goal error to steer action $\mathbf{a}$. \\

			\bottomrule
		\end{tabularx}
		\vspace{1em}
		\caption{The notation and symbols used to describe AICON's components.}
		\label{tab:aicon_components}
	\end{table}

	%----------------------------------------------------------------------
	\subsection{States -- Recursive Estimators}\label{sec:rec_est}
	%----------------------------------------------------------------------

	The first step is to identify the relevant states of the problem. These should capture all aspects of the world that lead to meaningfully different behavior.

	In our example, the task is to place a cube inside a drawer. The relevant states therefore include the poses of the cube, the drawer, and the robot's end-effector. In addition, the appropriate behavior depends on relational properties: whether the cube or drawer is grasped, and whether the cube is already inside the drawer. While these states are relatively obvious, additional states may become necessary depending on modeling assumptions. For instance, if the drawer pose is not known a priori, we must also estimate the parameters of its prismatic joint.

	For each state $\mathbf{x}$, we define a recursive estimator of the form
	\begin{equation}
		\mathrm{bel}(\mathbf{x}_{t}) = f(\mathrm{bel}(\mathbf{x}_{t-1}), \mathbf{z}_t, \mathbf{a}_t) \;,
	\end{equation}
	where $\mathrm{bel}(\mathbf{x}_t)$ denotes the belief over the state, $\mathbf{z}_t$ the measurement, and $\mathbf{a}_t$ the action, all at time $t$. These estimators typically correspond to variants of Bayesian filters, with representations tailored to the state. The design of each estimator follows a common procedure:
	\begin{enumerate}
		\item Choose an appropriate belief representation,
		\item Define a forward model describing state evolution (possibly dependent on actions $\mathbf{a}_t$),
		\item Incorporate measurement models for available observations $\mathbf{z}_t$ where possible.
	\end{enumerate}

	%------------------------------
	\subsubsection{Recursive Estimators in Our Example}
	%------------------------------

	For the continuous states, i.e., cube pose, drawer pose, end-effector pose, and prismatic joint parameters, we use multivariate Gaussian beliefs,
	\begin{equation}
		\mathbf{x}_\mathrm{cube}, \mathbf{x}_\mathrm{drawer}, \mathbf{x}_\mathrm{ee}, \mathbf{x}_\mathrm{pris} \sim \mathcal{N}(\boldsymbol{\mu}, \boldsymbol{\Sigma}) \;.
	\end{equation}

	For the discrete states, namely whether the cube or drawer is grasped and whether the cube is inside the drawer, we use categorical beliefs,
	\begin{equation}
		p_\text{cg}, p_\text{dg}, p_\text{cid} \in [0, 1] \;.
	\end{equation}

	Regarding forward models, many states (e.g., cube pose, drawer pose, prismatic joint, and categorical variables) are difficult to predict in isolation. For these, we assume a static model, $\mathbf{x}_{t+1} = \mathbf{x}_t$. For continuous states, this could be extended by including an estimated velocity. In contrast, the end-effector pose is directly influenced by control inputs, allowing a simple model:
	\begin{equation}
		\mathbf{x}_{t+1} = \mathbf{x}_t + \Delta t \; \mathbf{a}_{t} \;.
	\end{equation}

	Finally, some states can be directly observed: the end-effector pose via joint encoders, and grasp states via force sensing. Other states may lack direct measurements, motivating the introduction of constraints to enable information exchange between estimators.

	%----------------------------------------------------------------------
	\subsection{Constraints -- Active Interconnections}\label{sec:act_int}
	%----------------------------------------------------------------------

	While states define \emph{what} is represented, constraints define \emph{how states relate}. In contrast to classical planning approaches that assume perfect knowledge, AICON assumes only that states satisfy certain relationships. These constraints typically arise from physics, geometry, or domain knowledge.

	For example, the drawer pose and its prismatic joint are coupled through kinematics: the drawer must lie along its prismatic axis, and the axis must pass through the drawer. Such constraints allow information to flow between states, supporting both perception and action.

	To define constraints, we examine the identified states and determine their relationships. Each constraint is expressed as a function over the involved states,
	\begin{equation}
		c(\mathbf{x}_1, \mathbf{x}_2, \dots, \mathbf{x}_N) \overset{!}{=} 0 \;,
	\end{equation}
	which evaluates to zero if and only if the constraint is satisfied. Such an implicit form does not fix which states are inputs and which are outputs, yet it still defines local relationships between all connected states.\footnote{This follows from the implicit function theorem~\cite{lang2012fundamentalsImplicitFunction,jittorntrum1978implicit}. Wherever the Jacobian $\partial c / \partial \mathbf{x}_i$ is invertible, $c = 0$ locally defines $\mathbf{x}_i$ as a differentiable function of the remaining states, with derivatives $\partial \mathbf{x}_i / \partial \mathbf{x}_j = -(\partial c / \partial \mathbf{x}_i)^{-1}\, \partial c / \partial \mathbf{x}_j$. Since this holds for whichever state satisfies the invertibility condition, an implicit constraint defines local relationships and derivatives between connected states in either direction. This is what lets information and gradients flow through it without committing to an input--output direction in advance.} This formulation captures both equality and inequality constraints and facilitates inference. An inequality such as $x \le 5$, for instance, becomes
	\begin{equation}
		c(x) = \max(0,\, x - 5) \;.
	\end{equation}
	A satisfied inequality has $c = 0$ and $\partial c / \partial x = 0$, so it passes neither information nor gradient; it becomes ``active'' only when it is violated or at its boundary.

	%------------------------------
	\subsubsection{Active Interconnections in Our Example}
	%------------------------------

	In our example, several constraints arise:
	\begin{itemize}
		\item The drawer pose and prismatic joint are kinematically coupled:\begin{equation}
			c_\mathrm{kin}(\mathbf{x}_\mathrm{pris}, \mathbf{x}_\mathrm{drawer}) = 0\;.
		\end{equation}
		\item If the drawer or cube is grasped, its pose is constrained by the end-effector: \begin{equation}c_\mathrm{dg}(\mathbf{x}_\mathrm{ee}, \mathbf{x}_\mathrm{drawer}, p_\text{dg}) = 0 \;, \qquad\qquad c_\mathrm{cg}(\mathbf{x}_\mathrm{ee}, \mathbf{x}_\mathrm{cube}, p_\text{cg}) = 0\;.
		\end{equation}
		\item Whether the cube is inside the drawer depends on their poses and the drawer state: \begin{equation}
			c_\mathrm{cid}(\mathbf{x}_\mathrm{drawer}, \mathbf{x}_\mathrm{cube}, \mathbf{x}_\text{pris}, p_\text{cid}) = 0\;.
		\end{equation}
		\item Moreover, there are fairly simple constraints to how forces measured at the end-effector depend on whether something is grasped and how robot kinematics relate end-effector pose, the robot's joint encoder values, and the robot's actuation signals:
		\begin{equation}
			c_\mathrm{grasp}(\mathbf{s}_\mathrm{force}, p_\text{dg},  p_\text{cg}) = 0 \;, \qquad\qquad c_\mathrm{rob}(\mathbf{x}_\mathrm{ee}, \mathbf{s}_\mathrm{enc}, \mathbf{a}) = 0\;.
		\end{equation}
	\end{itemize}

\begin{figure}[t]
		\centering
		\begin{tikzpicture}
			\drawCubeDrawerNet{}
		\end{tikzpicture}
		\caption{Example AICON network for placing a cube into a drawer. The system is defined by states (e.g., poses and grasp probabilities) and constraints capturing kinematics, contacts, sensing, and actuation. Information flows through constraints during inference, allowing unobserved states to be estimated, while gradients from the goal propagate through the same structure to generate behavior.}\label{fig:example_network}
	\end{figure}
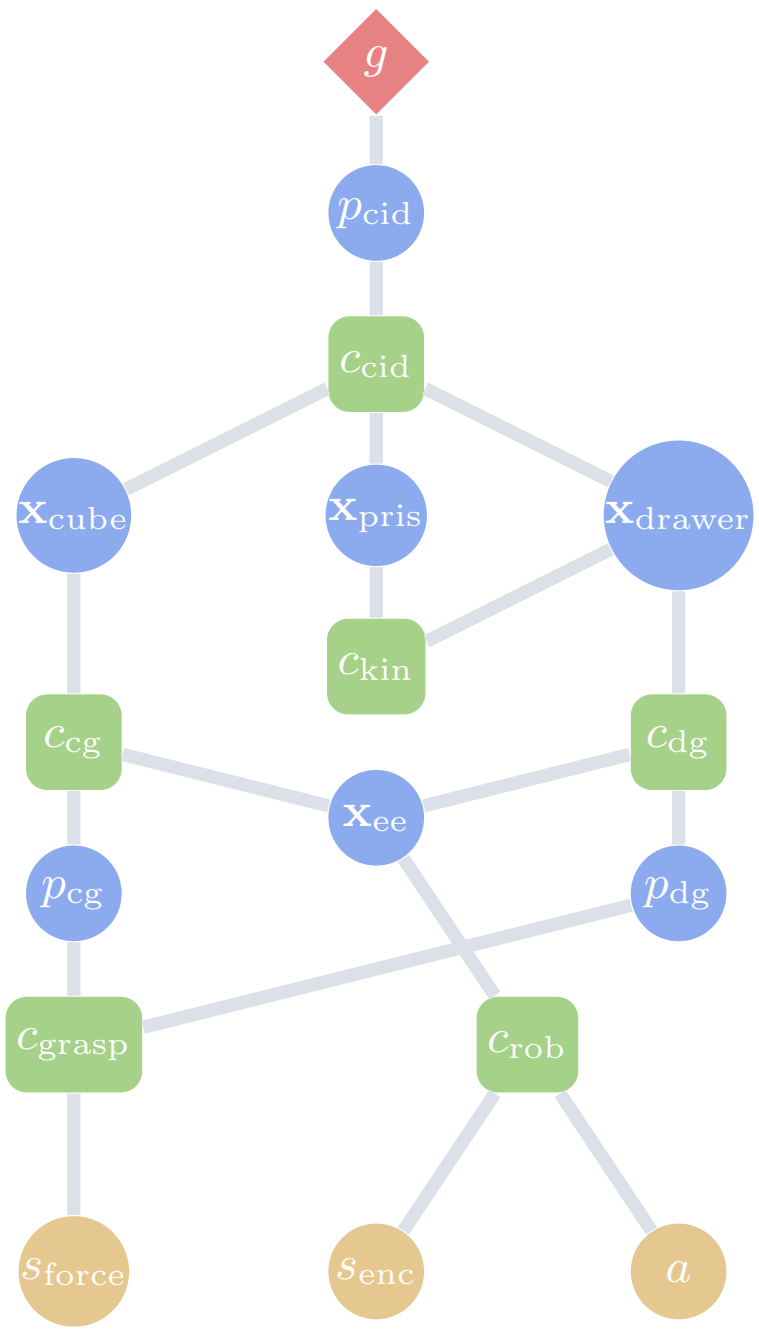

	In a real-world system, we might have additional constraints, for example based on projective geometry constraining how objects are seen through an in-hand camera, but these constraints suffice for our example. They form a network, shown in \Cref{fig:example_network}, through which information can propagate, enabling both state estimation and action selection.

	%----------------------------------------------------------------------
	\subsection{Passive Perception -- Constrained Probabilistic Inference}\label{sec:inference}
	%----------------------------------------------------------------------

	Given the network of estimators and constraints, perception emerges as a process of constrained belief updating. Each estimator maintains its belief while ensuring consistency with connected constraints.

	At each time step, an estimator:
	\begin{enumerate}
		\item Predicts its state via the forward model,
		\item Incorporates direct measurements,
		\item Updates its belief using information from constraints.
	\end{enumerate}
	Formally, for $M$ constraints,
	\begin{equation}
		\mathrm{bel}(\mathbf{x}_{t}) = p(\mathbf{x}_{t} | c_1(\bullet, \mathbf{x}_{t})) \dots p(\mathbf{x}_{t} | c_M(\bullet, \mathbf{x}_{t})) \; f(\mathrm{bel}(\mathbf{x}_{t-1}), \mathbf{z}_t, \mathbf{a}_t)\label{eq:constraint_inf}\;.
	\end{equation}

	\begin{skipbox}
		To implement \Cref{eq:constraint_inf} in practice, different inference schemes can be used depending on the chosen belief representations and structural assumptions. \Cref{tab:aicon_table} summarizes common combinations and their corresponding algorithms. These choices ensure that constrained probabilistic inference remains tractable while allowing information to be exchanged consistently across heterogeneous estimators.
	\end{skipbox}

	\begin{sidewaystable}
		\begin{skipbox}[unbreakable]
		\centering
		\small
		\setlength{\tabcolsep}{8pt}
		\def\arraystretch{1.5}
		\begin{tabular}{@{}>{\raggedright\arraybackslash}p{5.2cm}>{\raggedright\arraybackslash}p{3.9cm}>{\raggedright\arraybackslash}p{10.2cm}>{\raggedright\arraybackslash}p{2.2cm}@{}}	\toprule
			\textbf{Belief Representation} \newline \emph{Additional Assumptions} & \textbf{Algorithm} & \textbf{Central Equations} & \textbf{References} \\ \midrule

			All Gaussian, $x_i \sim \mathrm{Normal}(\mu_i,\Sigma_i)$ \newline \emph{independent uncertainties} &
			Kalman-style constraint update (KF/EKF/UKF depending on linearity) &
			Residual $c(x_1,\dots,x_N)=0$ with $H_i = \partial c / \partial x_i$. \newline
			Uncertainty over residual: $S = R + \sum_i H_i \Sigma_i H_i^\top$. \newline
			Gain: $K_i = \Sigma_i H_i^\top S^{-1}$. \newline
			Update: $\mu_i \leftarrow \mu_i - K_i\, c(\mu_1,\dots,\mu_N)$, \quad $\Sigma_i \leftarrow \Sigma_i - K_i H_i \Sigma_i$ &
			\cite{mengers2023combining,mengers2025noplan,martinmartin2022coupled,battaje2024information,mengers2024leveraging,pfisterer2025helping,mengers2026mechanistic,battaje2026building} \\

			All Gaussian, $x_i \sim \mathrm{Normal}(\mu_i,\Sigma_i)$ \newline \emph{strongly interdependent uncertainties} &
			Covariance Intersection &
			Instead of a gain-based update, perform Covariance Intersection between $\mathrm{Normal}(\mu_i,\Sigma_i)$ and \newline
			$\mathrm{Normal}\big(\mu_i - H_i^{\#} c(\mu_1,\dots,\mu_N),\ H_i^{\#} S (H_i^{\#})^\top\big)$, \newline
			where $H_i^{\#}$ is the pseudoinverse of $H_i$ &
			\cite{mengers2024leveraging} \\

			All categorical, $x_i \sim \mathrm{Cat}(x_i)$ &
			Finite summation (marginalizing) &
			For each value $X_i^k$ of $x_i$: \newline
			$p(x_i{=}X_i^k) \leftarrow p(x_i{=}X_i^k) \cdot \sum_{\{x_j: j\neq i\}} c(x_1,\dots,x_N)\prod_{j\neq i} p(x_j)$, \newline
			then renormalize &
			\cite{mengers2023combining,migacev2026planhumanreactiverobotics}\\

			Categorical $x_0 \sim \mathrm{Cat}(x_0)$ \& Gaussian $x_i \sim \mathrm{Normal}(\mu_i,\Sigma_i)$ \newline \emph{categorical ``activates'' constraint} &
			Modified Kalman update &
			Gaussian update as above, but gain becomes $K_i = p(x_0)\, \Sigma_i H_i^\top S^{-1}$. \newline
			Categorical update as above, using $S$ to determine the likelihood of the constraint given the Gaussian beliefs &
			\cite{mengers2025noplan,mengers2026mechanistic,battaje2026building} \\

			Categorical $x_0 \sim \mathrm{Cat}(x_0)$ \& Gaussian $x_i \sim \mathrm{Normal}(\mu_i,\Sigma_i)$ \newline \emph{categorical switches between constraint modes} &
			Switching KF &
			As above, but repeated for each value $X_0^k$ of $x_0$ with its respective constraint $c_k$. \newline
			Gain: $K_i^k = p(x_0{=}X_0^k)\,\Sigma_i (H_i^k)^\top (S^k)^{-1}$. \newline
			Combined as: $\mu_i \leftarrow \mu_i - \sum_k K_i^k\, c_k(\mu_1,\dots,\mu_N)$ &
			\cite{martinmartin2022coupled} \\

			All particles, $x_i \sim \{x_i^{[k]}, w_i^{[k]}\}_{k=1}^K$ \newline \emph{state space permits kernel interpolation} &
			Kernel density \& importance weighting &
			Evaluate constraint at each particle pair. \newline
			Reweight: $w_i^{[k]} \leftarrow w_i^{[k]} \cdot \kappa\!\left(c(x_1^{[k]},\dots,x_N^{[k]})\right)$ for kernel $\kappa$. \newline
			Then resample &
			\cite{lowin2026coupled} \\

			Particles $x_i \sim \{x_i^{[k]}, w_i^{[k]}\}_{k=1}^K$ \& Gaussian $x_j \sim \mathrm{Normal}(\mu_j,\Sigma_j)$ &
			Particles weighted; Gaussian approximates particles &
			Reweight particles as above using $\mu_j,\Sigma_j$. \newline
			Approximate particles as Gaussian: fit $\hat\mu_i, \hat\Sigma_i$ from $\{x_i^{[k]}, w_i^{[k]}\}$. \newline
			Then apply the Gaussian update to $x_j$ &
			-- \\

			Particles $x_i \sim \{x_i^{[k]}, w_i^{[k]}\}_{k=1}^K$ \& Categorical $x_0 \sim \mathrm{Cat}(x_0)$ &
			Particles weighted; categories weighted &
			Particle weights updated as above for each value $X_0^k$ of $x_0$. \newline
			Categorical updated as in the all-categorical case, \newline
			replacing the marginalization sum over $x_i$ with $\sum_k w_i^{[k]}\, c(\dots, x_i^{[k]}, \dots)$ &
			\cite{mengers2023combining,mengers2025scanpath} \\
			\bottomrule
		\end{tabular}
		\caption{Overview of algorithmic choices for message computation in constrained Bayesian filtering, depending on belief representations and structural assumptions.}
		\label{tab:aicon_table}
		\end{skipbox}
	\end{sidewaystable}

	%------------------------------
	\subsubsection{Perception in Our Example}
	%------------------------------

	To illustrate the resulting information flow, consider the constraint between the drawer pose and the prismatic joint.

	Initially, the parameters of the prismatic joint may be highly uncertain. In this case, little to no information flows from the joint estimate to the drawer pose through their shared constraint. However, once the drawer is observed to move, the system can begin to infer the parameters of the prismatic joint from this motion. As the estimate of the joint becomes more certain, it in turn constrains the drawer pose more strongly, since the drawer must lie along the estimated axis.

	This bidirectional exchange of information is characteristic of the entire network. Each estimator continuously updates its belief based on both local information and constraints, leading to a global, adaptive flow of information. Importantly, the direction and strength of this flow are not explicitly specified but emerge naturally from the relative uncertainties of the connected states.

	%----------------------------------------------------------------------
	\subsection{Behavior Generation -- Goals and Gradient Descent}\label{sec:actions}
	%----------------------------------------------------------------------

	Behavior generation is achieved by defining a scalar cost function $g(\mathbf{x})$ over the estimated states and propagating its gradient through the network to the actuators.

	For a given path $\mathrm{p}: g \leadsto \mathbf{a}$ through the network along a sequence of estimators $1, \dots, L$, the gradient is
	\begin{equation}
		\nabla^\mathrm{p}_\mathbf{a}\;g=\frac{\partial g(\mathbf{x}_1)}{\partial \mathbf{x}_1}\frac{\partial \mathbf{x}_1}{\partial \mathbf{x}_2} \dots \frac{\partial \mathbf{x}_{L}}{\partial \mathbf{a}} \label{eq:chain} \;.
	\end{equation}

	\Cref{eq:chain} contains only derivatives between states; the constraints do not appear explicitly. They are hidden in the factors: each factor $\partial \mathbf{x}_k / \partial \mathbf{x}_{k+1}$ is the local derivative that the constraint connecting $\mathbf{x}_k$ and $\mathbf{x}_{k+1}$ induces between them. This is how a constraint is \emph{read out for action}. For perception, the constraint residual drives a belief update of $\mathbf{x}_k$ (\Cref{tab:aicon_table}). For action, we ask how this update would change if $\mathbf{x}_{k+1}$ were different, that is, we differentiate the update with respect to the neighboring state. For the Gaussian update $\boldsymbol{\mu}_k \leftarrow \boldsymbol{\mu}_k - K_k\, c$, this gives
	\begin{equation}
		\frac{\partial \mathbf{x}_k}{\partial \mathbf{x}_{k+1}} = -K_k H_{k+1}\;, \qquad H_{k+1} = \frac{\partial c}{\partial \mathbf{x}_{k+1}} \label{eq:readout}\;.
	\end{equation}
	When $\mathbf{x}_k$ is much more uncertain than the other states in the constraint, the constraint is noise-free, and $H_k$ is invertible, the gain approaches $K_k \to H_k^{-1}$, and \Cref{eq:readout} reduces to the derivative given by the implicit function theorem (\Cref{sec:act_int}). In general, the derivative also reflects how confident the connected estimators are and, through the gain $K_k = p(x_0)\,\Sigma_k H_k^\top S^{-1}$ of a constraint activated by a categorical belief (\Cref{tab:aicon_table}), how strongly the system believes that the constraint currently holds. This last point is what makes sequencing possible, as we show below. Note that nothing here is designed as a gate: we only write down the regularity (for example, that a grasped object moves with the hand) and estimate whether it currently applies. That the derivative shrinks when the regularity is unlikely to hold is a consequence of the standard belief update, not an added mechanism.

	Rather than combining gradients from all possible gradient paths $\mathrm{p} \in \mathbb{P}$ to the action $\mathbf{a}$, which can lead to conflicts in action space, we select the steepest path $\mathrm{p}^*$ to update the action with gain $\mathbf{k}$:
	\begin{equation}
		\mathbf{a}_{t+1} = \mathbf{a}_{t} - \mathbf{k} \cdot \nabla^{\mathrm{p}^*}_\mathbf{a}\; g \;, \qquad \mathrm{p}^* = \argmax_{\mathrm{p} \in \mathbb{P}} \|\nabla^\mathrm{p}_\mathbf{a}\; g\| \label{eq:aicon_steepest}\;.
	\end{equation}

	As beliefs evolve, both gradients and the selected path change, resulting in adaptive and sequential behavior without explicitly encoding temporal logic. The switches between paths are not designed either: they follow from the changing gradient magnitudes and the steepest-path selection of \Cref{eq:aicon_steepest}.

	%------------------------------
	\subsubsection{Behavior Generation in Our Example}
	%------------------------------

	The goal is for the cube to be inside the drawer. Since this is already represented as a state, we define the cost as:
	\begin{equation}
		g(p_\text{cid}) = - p_\text{cid}\;.
	\end{equation}

\begin{figure}[tbp]
		\centering
		\begin{tikzpicture}[scale=0.9, every node/.append style={transform shape}]
			% (a) drawer closed, handle not grasped
			\begin{scope}[shift={(0,0)}]
				\drawCubeDrawerNet{A}
				\draw[aicon/gradient] (AG) -- (APcid);
				\draw[aicon/gradient] (APcid) -- (ACcid);
				\draw[aicon/gradient] (ACcid) -- (AXdrawer);
				\draw[aicon/gradient] (AXdrawer) -- (ACdg);
				\draw[aicon/gradient, aicon/short] ([xshift=4pt]ACdg.south) -- ([xshift=4pt]APdg.north);
				\draw[aicon/gradient, aicon/short] ([xshift=-4pt]APdg.north) -- ([xshift=-4pt]ACdg.south);
				\draw[aicon/gradient] (ACdg) -- (AXee);
				\draw[aicon/gradient] (AXee) -- (ACrob);
				\draw[aicon/gradient] (ACrob) -- (AA);
				\node[font=\large\bfseries, align=center] at (4.5,-1.2) {(a) Grasp handle};
			\end{scope}
			% (b) drawer grasped
			\begin{scope}[shift={(6,0)}]
				\drawCubeDrawerNet{B}
				\draw[aicon/gradient] (BG) -- (BPcid);
				\draw[aicon/gradient] (BPcid) -- (BCcid);
				\draw[aicon/gradient] (BCcid) -- (BXdrawer);
				\draw[aicon/gradient] (BXdrawer) -- (BCdg);
				\draw[aicon/gradient] (BCdg) -- (BXee);
				\draw[aicon/gradient] (BXee) -- (BCrob);
				\draw[aicon/gradient] (BCrob) -- (BA);
				\node[font=\large\bfseries, align=center] at (4.5,-1.2) {(b) Open drawer};
			\end{scope}
			% (c) drawer open, cube not grasped
			\begin{scope}[shift={(0,-10.4)}]
				\drawCubeDrawerNet{C}
				\draw[aicon/gradient] (CG) -- (CPcid);
				\draw[aicon/gradient] (CPcid) -- (CCcid);
				\draw[aicon/gradient] (CCcid) -- (CXcube);
				\draw[aicon/gradient] (CXcube) -- (CCcg);
				\draw[aicon/gradient, aicon/short] ([xshift=4pt]CCcg.south) -- ([xshift=4pt]CPcg.north);
				\draw[aicon/gradient, aicon/short] ([xshift=-4pt]CPcg.north) -- ([xshift=-4pt]CCcg.south);
				\draw[aicon/gradient] (CCcg) -- (CXee);
				\draw[aicon/gradient] (CXee) -- (CCrob);
				\draw[aicon/gradient] (CCrob) -- (CA);
				\node[font=\large\bfseries, align=center] at (4.5,-1.2) {(c) Grasp cube};
			\end{scope}
			% (d) cube grasped
			\begin{scope}[shift={(6,-10.4)}]
				\drawCubeDrawerNet{D}
				\draw[aicon/gradient] (DG) -- (DPcid);
				\draw[aicon/gradient] (DPcid) -- (DCcid);
				\draw[aicon/gradient] (DCcid) -- (DXcube);
				\draw[aicon/gradient] (DXcube) -- (DCcg);
				\draw[aicon/gradient] (DCcg) -- (DXee);
				\draw[aicon/gradient] (DXee) -- (DCrob);
				\draw[aicon/gradient] (DCrob) -- (DA);
				\node[font=\large\bfseries, align=center] at (4.5,-1.2) {(d) Carry cube};
			\end{scope}
		\end{tikzpicture}
		\caption{Emergent sequencing in the cube-in-drawer network of \Cref{fig:example_network}. Each panel shows the steepest gradient path in one phase of the task. (a) The drawer is closed, so only paths via the drawer have a non-zero gradient; since the handle is not yet grasped, the steepest one runs through the grasp belief $p_\mathrm{dg}$, and the robot reaches for the handle. (b) Once $p_\mathrm{dg}$ saturates, the path via $c_\mathrm{dg}$ takes over and the robot pulls the drawer open. (c) With the drawer open, paths via the cube become steep; the cube is not yet grasped, so the path through $p_\mathrm{cg}$ dominates. (d) Once the cube is grasped, the path via $c_\mathrm{cg}$ carries it into the drawer. No phase or transition is specified; the switch between paths follows from the current beliefs.}
		\label{fig:sequencing}
	\end{figure}
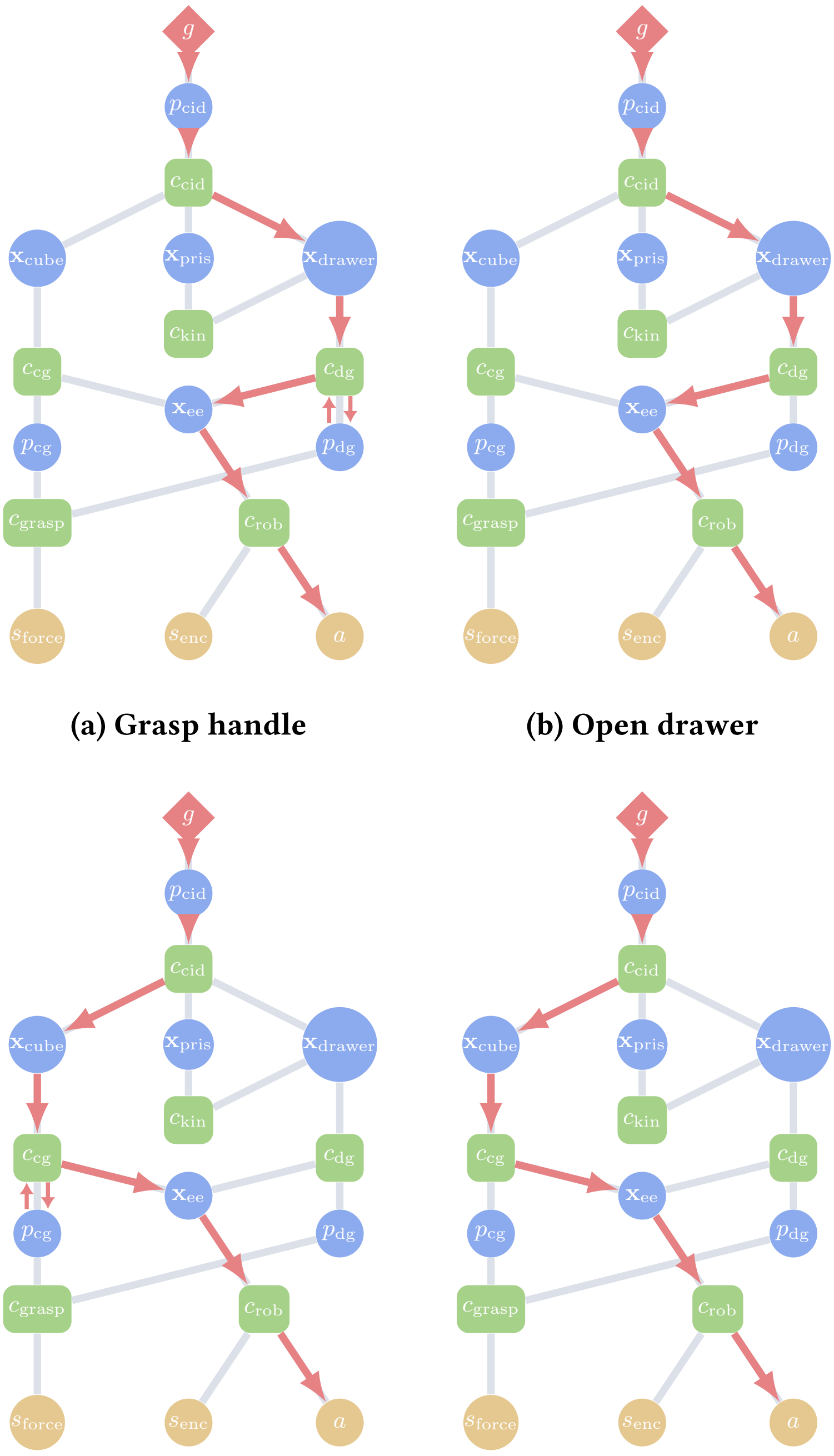

	In the network of \Cref{fig:example_network}, the action $\mathbf{a}$ influences the end-effector pose $\mathbf{x}_\mathrm{ee}$ through $c_\mathrm{rob}$, and every gradient path to the goal must therefore pass through $\mathbf{x}_\mathrm{ee}$. From there, it can reach $p_\mathrm{cid}$ either via the drawer ($c_\mathrm{dg}$) or via the cube ($c_\mathrm{cg}$). \Cref{fig:sequencing} shows the path that is steepest in each phase of the task. Let us walk through how the current beliefs shape the steepness of these paths. We call this effect \emph{gating}, but it is not a mechanism we add to the graph: it falls out of differentiating the belief updates and selecting the steepest path.

	\paragraph{The grasp belief shapes the gradient paths.}
	Consider the path via the cube,
	\begin{equation}
		\nabla^{\mathrm{carry}}_\mathbf{a}\, g = \frac{\partial g}{\partial p_\mathrm{cid}}\,\frac{\partial p_\mathrm{cid}}{\partial \mathbf{x}_\mathrm{cube}}\,\frac{\partial \mathbf{x}_\mathrm{cube}}{\partial \mathbf{x}_\mathrm{ee}}\,\frac{\partial \mathbf{x}_\mathrm{ee}}{\partial \mathbf{a}}\;.
	\end{equation}
	The factor $\partial \mathbf{x}_\mathrm{cube} / \partial \mathbf{x}_\mathrm{ee}$ is read out from $c_\mathrm{cg}$ via \Cref{eq:readout}. The regularity encoded in $c_\mathrm{cg}$, that a grasped cube moves with the hand, holds only if the cube is grasped. Its update is therefore weighted by the grasp belief, its gain contains $p_\mathrm{cg}$, and so $\|\nabla^{\mathrm{carry}}_\mathbf{a} g\| \propto p_\mathrm{cg}$. Intuitively, moving the hand only moves the cube if the cube is held. While the cube lies on the table, $p_\mathrm{cg} \approx 0$ and this path is practically closed.

	A second path runs through the grasp belief itself,
	\begin{equation}
		\nabla^{\mathrm{grasp}}_\mathbf{a}\, g = \frac{\partial g}{\partial p_\mathrm{cid}}\,\frac{\partial p_\mathrm{cid}}{\partial \mathbf{x}_\mathrm{cube}}\,\frac{\partial \mathbf{x}_\mathrm{cube}}{\partial p_\mathrm{cg}}\,\frac{\partial p_\mathrm{cg}}{\partial \mathbf{x}_\mathrm{ee}}\,\frac{\partial \mathbf{x}_\mathrm{ee}}{\partial \mathbf{a}}\;.
	\end{equation}
	Here, $\partial \mathbf{x}_\mathrm{cube} / \partial p_\mathrm{cg}$ expresses that a higher grasp belief would bind the cube more strongly to the hand, and $\partial p_\mathrm{cg} / \partial \mathbf{x}_\mathrm{ee}$ expresses that the grasp belief, updated through $c_\mathrm{cg}$ (and, in a real system, through gripper closing and force sensing via $c_\mathrm{grasp}$), rises as the hand reaches the cube. Schematically, this path is steep while the grasp belief can still be raised and flattens once it saturates, whereas the carry path does the opposite. Following the steepest path therefore produces ``approach and grasp the cube'' first and ``carry the cube'' second (\Cref{fig:sequencing}c,d). Neither was specified as a subgoal, and no switching condition was designed; both emerge from how the steepness of the two paths changes with $p_\mathrm{cg}$, together with steepest-path selection. For a complete treatment of a system built this way, see \citet{mengers2025noplan}.

	\paragraph{The drawer comes first if it is closed.}
	If the drawer is closed, the cube cannot be inside it, no matter where the cube is. Then $\partial p_\mathrm{cid} / \partial \mathbf{x}_\mathrm{cube} \approx 0$, and both cube paths vanish, while the derivative of $p_\mathrm{cid}$ with respect to the drawer opening is large. The steepest path now leads via the drawer, and the same effect arises from $p_\mathrm{dg}$: first, the path through the grasp belief makes the robot grasp the handle (\Cref{fig:sequencing}a); once $p_\mathrm{dg}$ saturates, the path via $c_\mathrm{dg}$ makes it pull the drawer open (\Cref{fig:sequencing}b). As the drawer opens, $\partial p_\mathrm{cid} / \partial \mathbf{x}_\mathrm{cube}$ grows, the cube paths take over, and the behavior continues as described above.

	Importantly, these behaviors are not pre-programmed sequences. They emerge from the current state of the system: as the state changes, the system continuously reselects the most promising direction toward the goal. If the cube slips from the hand, $p_\mathrm{cg}$ drops, the carry path closes, and the grasp path becomes steepest again; if someone closes the drawer, $\partial p_\mathrm{cid} / \partial \mathbf{x}_\mathrm{cube}$ vanishes and the drawer paths take over. These are exactly the recovery transitions that had to be added by hand to the state machine in \Cref{fig:state_machine}.

	%----------------------------------------------------------------------
	\subsection{Leveraging Nullspaces to Resolve Conflicting Goals}\label{sec:nullspace}
	%----------------------------------------------------------------------

	So far, AICON selects the single steepest gradient at each moment. This works well when subgoals are compatible or arise sequentially, but fails when multiple objectives are simultaneously active and their gradients conflict, pointing in opposing directions in action space. A robot navigating around an obstacle while pursuing a goal behind it is a canonical example: the attractive and repulsive gradients may directly oppose each other, trapping myopic gradient descent in a local minimum.

	AICON addresses this by extending its gradient-based action selection with projections and explorations in the nullspace~\cite{mengers2026riding,dietrich_overview_2015}. Rather than selecting one gradient, the system projects lower-priority gradients into the nullspace of higher-priority ones---the subspace of directions that do not worsen the higher-priority objective. Crucially, this happens \emph{where} the gradients meet: at the first quantity in the graph whose state both goals seek to change, not only in action space. Once the conflict is resolved there, the combined gradient propagates onward as a single gradient and is projected again only where it meets gradients of further goals. Resolving conflicts only further downstream, for example in joint space, arbitrates between projections of a conflict whose structure is no longer visible. On a real robot pushing task, restricting projection to joint space lowered success from 98\% to 12\%~\cite{mengers2026riding}. Priorities, in turn, are not fixed in advance but read continuously from the current gradient magnitudes, so the hierarchy adapts \emph{when} the interaction structure changes. This resolves many conflicts before local minima form, and when irresolvable conflicts remain, the nullspace provides a direction for exploration along the obstacle or constraint boundary until the interaction structure changes and normal goal pursuit can resume.

	Returning to \Cref{fig:intro:aicon_vs_others}, nullspace projections extend AICON's capabilities further into the top-right quadrant: the class of tasks solvable without planning now includes not only sequential subgoals but also concurrent conflicting objectives, as long as the conflicts arise from currently active interactions rather than from long-horizon dependencies of the kind found in Tower of Hanoi or a Rubik's Cube. The limits of this approach are therefore not arbitrary but reflect a principled boundary: conflicts that only emerge through future state sequences remain genuinely hard, while conflicts between simultaneously relevant goals can be resolved reactively~\cite{mengers2026riding}.

	%------------------------------
	\subsubsection{Nullspace Projections in Our Example}
	%------------------------------

	In the cube-in-drawer task, conflicting gradients arise naturally as soon as we add an obstacle between the robot and the cube. The gradient pulling the end-effector toward the cube and the gradient pushing it away from the obstacle may point in nearly opposite directions in action space, trapping myopic gradient descent in a local minimum.

	With nullspace projections, the system resolves this conflict continuously. The dominant gradient---toward the cube---takes priority, and the obstacle avoidance gradient is projected into its nullspace, allowing the robot to make progress toward the cube while also avoiding the obstacle. Both gradients reach the action through the end-effector pose $\mathbf{x}_\mathrm{ee}$, so this is where they meet and where the projection takes place. When the two gradients are nearly opposing and no such projection resolves the conflict, the robot enters brief nullspace exploration, moving along the obstacle boundary until the interaction structure shifts and a direct path toward the cube opens.

	%======================================================================
	\section{How to Build an AICON Graph for a New Problem}\label{sec:howto}
	%======================================================================

	Building an AICON graph is less like programming and more like modeling: you are describing the structure of your world, not writing procedures for how to act in it. The process is inherently iterative---you start with a minimal model, deploy it, observe where it fails or behaves unexpectedly, and refine it by identifying which regularity is missing, incomplete, or incorrectly captured. This section walks through the four recurring steps of that process. For a hands-on walk-through of the same steps in code, see the tutorial in the AICON repository (\url{https://github.com/tu-rbo/aicon}).

	%----------------------------------------------------------------------
	\subsection*{Step 1: Identify the Relevant States}
	%----------------------------------------------------------------------

	Start by asking what quantities matter for the task or behavior at hand---either because they are directly task-relevant, or because they influence how other relevant quantities interact. Each identified quantity becomes a recursive estimator. At this stage, err on the side of including states that seem plausibly relevant; unnecessary ones can be removed later, and missing ones are a common source of failure.

	For each estimator, three design choices are required:
	\begin{enumerate}
		\item A \emph{belief representation} suited to the quantity (Gaussian, categorical, particles---see \Cref{tab:aicon_table}),
		\item A \emph{forward model} capturing how the quantity evolves over time, possibly depending on actions,
		\item A \emph{measurement model} wherever direct observations are available.
	\end{enumerate}

	%----------------------------------------------------------------------
	\subsection*{Step 2: Identify the Interdependencies}
	%----------------------------------------------------------------------

	Quantities in the world are rarely independent. Ask how the quantities you have identified relate to each other---through physics, geometry, kinematics, or domain knowledge. Each such relationship becomes an active interconnection: a differentiable constraint function that allows information to flow between connected estimators (\Cref{sec:act_int}).

	A useful heuristic is to look for cases where knowing one quantity well would help you estimate another, or where two quantities are coupled through a known physical law. Incomplete or missing interconnections are the most common source of poor performance in practice.

	%----------------------------------------------------------------------
	\subsection*{Step 3: Define the Goal}
	%----------------------------------------------------------------------

	Express what you want to achieve as a differentiable scalar cost function over one or more estimated states. The goal does not need to directly reference low-level quantities---it can be defined over high-level relational states, such as whether an object is grasped or whether two poses are aligned. The gradient of this cost propagates through the network via the chain rule (\Cref{eq:chain}), so the goal only needs to be defined at one point in the graph.

	%----------------------------------------------------------------------
	\subsection*{Step 4: Identify What You Can Control}
	%----------------------------------------------------------------------

	Determine which quantities are directly actionable---typically velocities or forces at actuators. These enter the graph as actuator nodes, connected to the states they influence through constraints, just like sensors. In \Cref{fig:example_network}, for example, the action $\mathbf{a}$ enters through $c_\mathrm{rob}$, which relates it to the end-effector pose and the joint encoders. States that the action drives directly, such as $\mathbf{x}_\mathrm{ee}$, additionally use $\mathbf{a}_t$ in their forward model (\Cref{sec:rec_est}). The gradient from the goal to the actuators is then computed via backpropagation through the graph, yielding the action at each step (\Cref{eq:aicon_steepest}).

	%----------------------------------------------------------------------
	\subsection*{Iterating on the Model}
	%----------------------------------------------------------------------

	After deploying an initial graph, failures typically fall into a small number of recognizable categories. The system gets stuck because a necessary gradient path does not exist---indicating a missing interconnection or state. Behavior is erratic because a constraint is too tight or too loose---indicating an incorrectly specified interconnection. The system acts on the wrong quantity because the goal is placed at the wrong point in the graph.

	In each case, the fix follows the same logic: identify which regularity is absent or incorrectly captured, and add or correct the corresponding component. Because AICON's components are modular, changes are local---adding a new estimator or interconnection does not require redesigning the rest of the graph.

	%======================================================================
	\section{Interfacing AICON with Other Approaches}\label{sec:interfacing}
	%======================================================================
	AICON's interconnected recursive estimators provide a world representation that generates robust behavior on its own. This representation can also serve as an interface for other methods, from high-level planners to low-level controllers and learning paradigms.

	\Cref{sec:intro_examples} contrasted AICON with planning and control but did not discuss how they can be combined. Interfacing AICON with a high-level planner, for instance to set AICON's goals in tasks that lie beyond its structural ceiling (\Cref{sec:lim_ceiling}), is a natural direction that we do not cover here. Instead, we explain how AICON interfaces with learning paradigms such as reinforcement learning (RL) and learning from demonstration (LfD). This interface benefits both sides: AICON's inductive biases support learning, while learning methods enable structured adaptation within AICON. We outline one specific example for each learning paradigm, although other examples are possible.

	%----------------------------------------------------------------------
	\subsection{Reinforcement Learning}\label{sec:rl}
	%----------------------------------------------------------------------

	Reinforcement learning (RL) specifies goals through reward functions, but it often suffers from poor sample efficiency. In standard end-to-end formulations, the policy must infer not only the task objective but also the relevant physical regularities, including kinematics, geometry, and sensorimotor constraints. This makes training expensive and often limits transfer between domains, embodiments, and object configurations.

	AICON mitigates this problem by giving the policy a structured head start. AICON already encodes task-invariant world regularities through recursive estimators and active interconnections. As a result, the RL policy does not need to rediscover rigid-body physics or kinematic constraints through trial and error. Instead, it receives AICON state estimates and candidate gradient paths that are already grounded in the world's regularities. These gradient paths provide a strong mathematical prior: rather than mapping raw observations directly to actions, the policy can choose among feasible, world-consistent directions toward the goal or its relevant subgoals.

	This interface yields a compact and sample-efficient policy. At each time step, AICON provides the current state estimate and a set of gradient paths. The RL policy then selects, weights, or modulates these paths instead of producing low-level motor commands from scratch. The resulting action space is therefore structured by AICON's graph: all available gradient paths are feasible directions for reducing the task cost, but they may differ in their long-term utility. A policy can be trained with standard RL algorithms, such as proximal policy optimization (PPO), to learn which path should be followed in each context.

	This design also addresses a limitation of AICON's default action-selection rule. Without a learned policy, AICON selects the steepest available gradient. This rule is effective from AICON's local perspective because every candidate path encodes a feasible way to make progress toward the goal. However, steepest descent cannot encode preferences over the solution trajectory, nor can it anticipate that a less direct path may lead to a better state later. RL complements AICON by learning these long-horizon preferences from reward. By rewarding successful path selections over time, RL can learn when to deviate from the locally steepest gradient in favor of a path with higher long-term utility.

	The policy can also be designed to handle a variable number of gradient paths. For example, an attention-based architecture can process the set of available paths without assuming a fixed ordering or fixed cardinality. This allows the same policy to scale beyond the training regime: if new objects, targets, robots, or constraints appear, AICON can add the corresponding estimators and interconnections to the graph, thereby exposing new gradient paths to the policy. The policy can then evaluate these paths through the same interface, rather than requiring a new output structure or complete retraining.

	AICON's modularity further supports transfer. When the robot embodiment, object geometry, sensor configuration, or environment changes, the corresponding components in the AICON graph can be replaced or extended. The learned RL policy can remain unchanged as long as the state estimates and gradient-path interface stay within the policy's input distribution. This makes few-shot or zero-shot sim-to-real transfer possible, even under substantial distribution shifts such as changes in robot morphology or object type. In this way, the problem factors into a world part and a task part~\cite{sebastian2026world}: AICON supplies the transferable world model, while RL learns task-specific preferences over the feasible solution paths that this world model exposes. We have already implemented this interface successfully for multiple tasks in multi-robot navigation and bi-manual manipulation~\cite{sebastian2026world}.

	%----------------------------------------------------------------------
	\subsection{Learning from Demonstration}\label{sec:lfd}
	%----------------------------------------------------------------------

	Learning from Demonstration (LfD) leverages a denser learning signal than RL because expert trajectories specify successful actions at every time step rather than only delayed feedback through rewards. However, LfD is prone to overfitting and often fails when the environment deviates from the demonstration data, i.e., in out-of-distribution scenarios.

	AICON provides a natural scaffold to address this limitation. Instead of using demonstrations only to train an end-to-end policy, we can use them to learn or refine individual components of the AICON graph. A particularly relevant example is the learning of active interconnections. These interconnections encode differentiable relationships between estimated world quantities, such as geometric constraints, contact conditions, visibility relations, or kinematic couplings. Some of these relationships are straightforward to specify by hand, whereas others may be difficult to model explicitly because they depend on complex geometry, changing contact regimes, or task-specific interaction patterns.

	Suppose that the structure of the AICON graph is known, but one active interconnection is unknown or only approximately specified. We can parameterize this interconnection as a differentiable function and learn its parameters from demonstrations. The surrounding graph then constrains the learning problem. Neighboring recursive estimators provide structured state estimates, known interconnections constrain the evaluation of these estimates, and the specified goal cost defines the gradient propagated through the graph. The learned interconnection is therefore not fit in isolation; it must remain consistent with the local physical regularities already encoded in AICON.

	Demonstrations provide an additional source of supervision through the observed actions. In AICON, actions arise from gradients propagated from the goal cost through recursive estimators and active interconnections to the actuators. Thus, a demonstrated action can be interpreted as evidence about the gradient that the learned interconnection should produce. Learning can therefore minimize not only the discrepancy between predicted and observed states, but also the discrepancy between the action predicted by AICON's gradient flow and the action taken by the demonstrator. In this sense, demonstrations provide a derivative-level learning signal: they supervise how the learned interconnection should shape the gradient field, not only what state it should predict.

	This is analogous in spirit to physics-informed learning. The learned component is flexible enough to capture relationships that are difficult to specify analytically, but it is regularized by the known AICON graph and by the requirement that its gradients explain the demonstrated behavior. As a result, LfD can be used not only to select among existing gradient paths, as in the RL interface, but also to improve the world model itself, e.g., by learning active interconnections. This enables AICON to incorporate task-relevant regularities from demonstrations while preserving the compositional structure and physical grounding that support generalization and robustness.

	%======================================================================
	\section{Relationship to Other Approaches}
	%======================================================================
	\label{sec:relationship}

	AICON shares similarities with many established methods. To clarify how it
	relates to each, we organize the comparison by level of specificity. We first
	situate AICON within broad \emph{general formalisms}---mathematical frameworks
	that define classes of possible systems. We then compare AICON to \emph{specific methods} that share concrete implementation details with it and highlight the differences.

	%----------------------------------------------------------------------
	\subsection{General Formalisms}\label{sec:formalisms}
	%----------------------------------------------------------------------

	%------------------------------
	\subsubsection{Probabilistic Graphical Models}
	%------------------------------

	AICON represents the world as a set of states (or belief states) and constraints between them. This resembles a large body of work in probabilistic graphical models (PGMs), which includes numerous variants: factor graphs, Bayesian networks, Markov random fields, among others. To situate AICON within this landscape, we disentangle three aspects: (1) how the distribution is represented structurally, (2) how much cross-correlation between variables is tracked, and (3) how inference is performed.

	\paragraph{Representation.}
	AICON's graph is bipartite: both states and constraints are explicit nodes, and edges only connect states to constraints. This matches the structure of a factor graph with variable nodes and factor nodes. It differs from Bayesian networks and Markov random fields (MRF), where only variables are nodes. AICON's graph is also undirected, like factor graphs and MRFs: a constraint $c(x_1, x_2) = 0$ does not assign a causal direction between the variables. Which variable informs which is determined at runtime by relative uncertainties, not by graph structure. This contrasts with Bayesian networks, where directed edges encode a fixed causal relationship.

	\paragraph{Cross-correlation.}
	Where does AICON sit on the spectrum of how much joint structure to track? At
	one extreme, a \emph{monolithic Bayes filter} over all quantities would
	correctly capture every cross-correlation but is computationally intractable
	and requires restructuring whenever a new quantity is added. At the other
	extreme, a \emph{mean-field approximation} treats all quantities as
	independent, discarding all correlations. AICON occupies a middle ground: it
	does not track the full joint, but its connections selectively maintain the
	dependencies that matter most---gaining modularity and tractability while
	preserving the relationships that drive useful behavior. The price of this middle ground is discussed in \Cref{sec:lim_overconfidence}.

	\paragraph{Inference.}
	Though AICON shares representational similarity with factor graphs, its inference mechanism differs substantially. Classical factor graph algorithms (belief propagation, expectation propagation) iterate messages to convergence, seeking to recover a global joint distribution or its marginals. In AICON, each quantity maintains its own local belief via a recursive estimator, connections pass a single message per timestep without iteration, and the graph can grow over time with old nodes never revisited.

	Consequently, AICON is best understood as a real-time, constraint-augmented, distributed Bayesian filtering architecture that borrows the representational structure of factor graphs but not their inference semantics.

	%------------------------------
	\subsubsection{Free Energy Principle and Active Inference}
	%------------------------------

	The Free Energy Principle (FEP) states that an agent perceives and acts in the world by minimizing variational free energy $\mathcal{F}$, defined as
	\begin{equation}
		\mathcal{F} = \mathbb{E}_{q_{\phi}}\!\left[\log q_{\phi}(x|\mathit{data}) - \log p_{\theta}(x,\mathit{data})\right] = - \log p_{\theta}(\mathit{data}) + D_{\mathrm{KL}}\!\left(q_{\phi}(x|\mathit{data})\,\|\, p_{\theta}(x|\mathit{data})\right),
	\end{equation}
	where $x$ are the environmental states, $\mathit{data}$ is the sensory input, $p_\theta$ is the agent's internal generative model of the world, and $q_\phi$ is the inference or recognition model. Since the KL divergence is non-negative, $\mathcal{F}$ is an upper bound on the surprise $-\log p_\theta(\mathit{data})$.
	Minimizing $\mathcal{F}$ adjusts the parameters $\theta$ and $\phi$, thus improving the generative model and the inference model.
	This covers not only perception but also action: the agent can choose actions that bring about the sensory input it expects under its prior preferences.

	FEP is a broad framework that integrates perception and action but does not specify how such a system should be implemented, e.g., Kalman filters and many ML methods can be derived from the FEP. As a result, AICON can also be viewed as an instantiation of FEP or Active Inference.
	The \emph{prior preferences} of Active Inference (preferred observations or states, sometimes called the agent's ``desires'') map onto our goal function. Using a goal function, we specify how a state should be and choose appropriate actions that reduce the cost and perceptual uncertainty.

	AICON shares FEP's core philosophy: perception and action are driven by a
	common objective. But AICON departs from the standard Active Inference
	mechanism in two ways. First, it does not maintain a single generative model;
	inference is distributed across local estimators coupled through connections.
	Second, AICON's action selection is purely myopic---it descends the goal
	gradient given current state estimates, without predicting future
	observations. Active Inference selects actions by evaluating their expected
	consequences; AICON selects actions by evaluating their instantaneous gradient
	magnitude.

	%------------------------------
	\subsubsection{Energy-Based Models}
	%------------------------------

	Energy-Based Models (EBMs) assign a scalar energy to a configuration of
	variables; valid configurations have low energy. Inference amounts to gradient
	descent on the energy with respect to all variables simultaneously, seeking a
	joint low-energy state. AICON's goal function plays the role of the energy,
	and action selection is gradient descent on this goal. Therefore, AICON's action selection shares some conceptual similarity with EBMs.

	However, there are some key differences. In a standard EBM, all variables are optimized jointly. In AICON,
	only the action variables are updated by descending the goal gradient. State
	variables are updated by local recursive estimators (Bayesian filters), not by minimizing
	a single global energy function. Another key difference concerns which gradient is followed. AICON's total gradient of the goal with respect to the action, the sum over all paths, is unique, just like the gradient of an EBM. But AICON does not follow it: it decomposes the total gradient into contributions from different paths and selects one of them (\Cref{eq:aicon_steepest}), whereas an EBM follows the full gradient. Last, the energy function in EBMs is fixed during inference, while AICON's goal gradient landscape is state dependent.

	%------------------------------
	\subsubsection{Markov Decision Processes (MDPs) and Partially Observable Markov Decision Processes (POMDPs)}
	%------------------------------
	Markov Decision Processes (MDPs) and their partially observable extension (POMDPs) are the canonical mathematical framework for sequential decision-making under uncertainty~\cite{puterman1994markov,kaelbling1998planning}. They are deeply related to planning: both formalize the problem of selecting actions to maximize long-term reward, and both require a known transition model of the world to do so. An MDP assumes the current state is fully observable; a POMDP relaxes this by maintaining a belief over hidden states, but still requires an accurate generative model of how observations and states relate.

	The appeal of these frameworks is precise and real: under their assumptions, they can guarantee mathematical properties such as optimality of the resulting policy. AICON makes no such guarantees. It selects the locally steepest gradient rather than a globally optimal action sequence, and its world model is an approximation that is updated continuously rather than specified in advance.

	The limitation of MDP and POMDP methods, however, is also precisely their strength: they depend entirely on the accuracy of their models and on the feasibility of searching the resulting state or belief space. For real-world problems, both conditions are hard to meet. Transition models that correctly capture physics, contact, and perception uncertainty are notoriously difficult to obtain. The belief space of a POMDP grows exponentially with the planning horizon, making exact solutions intractable except in small, well-specified domains. And crucially, anticipating future observations correctly over long horizons---which POMDP planning requires---becomes increasingly unreliable as model errors compound.

	AICON trades these guarantees for a different kind of robustness: it does not anticipate future states but responds continuously to the current one, and its model is grounded in real sensor measurements rather than a fixed prior. This makes it difficult to provide formal optimality guarantees, but it also means the system degrades gracefully under the model errors and disturbances that are unavoidable in the real world. For POMDP-based methods, achieving such graceful degradation outside controlled settings remains challenging, because both the models and the belief-space approximations they rely on are hardest to get right there~\cite{kaelbling1998planning}.

	%----------------------------------------------------------------------
	\subsection{Specific Methods}\label{sec:specific}
	%----------------------------------------------------------------------

	%----------------------------------------------------------------------
	\subsubsection{Model Predictive Control and Closed-Loop Replanning}\label{sec:mpc}
	%----------------------------------------------------------------------

	Model predictive control (MPC) optimizes an action sequence over a finite horizon using a forward model, executes only the first action, re-senses, and repeats. Closed-loop replanning does the same at the level of task or motion plans: it re-perceives the world, typically when execution deviates from the plan, and computes a new plan from the current state. Both methods re-ground themselves in the large world every cycle. By the criteria of \Cref{fig:intro:aicon_vs_others}, they therefore compete with AICON for the top right quadrant, and it is worth being precise about what distinguishes them.

	At first sight, MPC with a horizon of a single step looks almost identical to AICON: both select the next action by gradient-based optimization of a one-step objective. The substantive difference is structural~\cite{battaje2026building}. MPC optimizes over a single, unified forward model. Even when it combines several models or constraints, how they interact is fixed at design time or scheduled by hand, a \emph{static composition}. AICON's gradient instead composes contributions from many paths through its graph, and the influence of each regularity on the action adapts online to how informative it currently is: through the relative uncertainties of the connected estimators and through the beliefs over whether constraints currently hold (\Cref{eq:readout}). In this sense, AICON's composition of regularities is \emph{adaptive}. To obtain comparable context sensitivity from a unified MPC model, one would need either explicit switching logic or prediction horizons long enough for the optimizer to discover the different regimes by itself~\cite{battaje2026building}.

	Both remedies have a price. Switching logic brings back the hand-specified transitions of \Cref{fig:state_machine}. Long horizons require rollouts: MPC and replanning predict the consequences of candidate action sequences with a forward model and search over them. Within each horizon, they trust their small world model, their prediction errors compound, and their computational cost grows with the horizon length and the dimensionality of the action space. AICON performs no rollouts; it only evaluates local derivatives of its current beliefs (\Cref{eq:chain,eq:readout}).

	This also determines where the subgoal structure comes from. In MPC, subgoals appear only if the horizon is long enough, and the cost shaped well enough, for the optimizer to discover that, say, grasping the handle is a prerequisite for placing the cube. Closed-loop replanning obtains subgoals from search over a symbolic or geometric domain model. In AICON, the subgoal structure is read off the graph: which subgoal is pursued next follows from which gradient path is currently open, as in \Cref{fig:sequencing}, independently of any horizon.

	A further difference concerns state. MPC typically takes the current state as given, either measured directly or provided by a separate estimator. In AICON, estimation is part of the same graph that generates behavior: recursive estimators retain history in their beliefs and exchange information across modalities, which helps resolve perceptual aliasing~\cite{battaje2026building}, and these same beliefs determine which gradient paths are open.

	The differences cut both ways. MPC and replanning can anticipate dependencies that lie within their horizon but are not visible in the current gradient, which is exactly what AICON cannot do (\Cref{sec:lim_ceiling}). AICON, in turn, avoids the computational cost of search and the compounding of model errors over the horizon, and its sequencing is not bounded by a horizon length.

	%----------------------------------------------------------------------
	\subsubsection{Potential Fields}
	%----------------------------------------------------------------------

	Potential field methods are the most direct ancestor of AICON's action
	selection: define a scalar cost over the robot's state and descend its
	gradient. The classical limitation is well known: static potential fields
	suffer from local minima that trap the robot. AICON addresses this not by designing a minimum-free potential (as in
	navigation functions), but by dynamically recomposing the potential itself.

	%----------------------------------------------------------------------
	\subsubsection{Behavior-based Robotics}
	%----------------------------------------------------------------------

	Behavior-based robotics (BBR) approaches generate complex reactive behavior by selecting behaviors from a behavioral repertoire. These repertoires contain basic behaviors, such as obstacle avoidance, and can be extended to include more complex behaviors. BBR then runs one or multiple behaviors based on incoming sensor information without modeling or planning.

	AICON uses another idea to decompose the task of behavior generation. While BBR decomposes complex behavior into simpler ``behavior primitives'', AICON does none of that. Instead, behavior emerges from the task and the world's encoded regularities. From this, behaviors such as error recovery and interactive perception emerge, which have to be encoded explicitly into the BBR's behavioral repertoire.

	%----------------------------------------------------------------------
	\subsubsection{Riemannian Motion Policies and Nullspace Control}
	%----------------------------------------------------------------------

	Like nullspace control and Riemannian Motion Policies (RMPs), AICON resolves multi-objective behavior by composing local gradient-based controllers, retaining the robustness and reactivity of control. Traditional nullspace control and RMPs, however, rely on fixed, hand-designed hierarchies or metrics that determine how objectives interact, effectively hard-coding task structure and limiting adaptability. In contrast, AICON treats goal interactions as state-dependent: gradients not only drive behavior but are dynamically modulated by the information flow between components, yielding an adaptive hierarchy that emerges from the current world state. By projecting gradients into one another's nullspaces wherever they meet in the graph, rather than only in action or configuration space, while continuously reshaping their relative influence (\Cref{sec:nullspace}), AICON prevents and escapes local minima caused by conflicting objectives without predefining priorities, subgoals, or switching logic. This enables complex, multi-objective behavior with the robustness of myopic control, while avoiding the rigidity of classical nullspace schemes and the fixed geometric assumptions of RMPs.

	%----------------------------------------------------------------------
	\subsubsection{World Models}
	%----------------------------------------------------------------------

	In recent years, \emph{world models} have progressed from classical control representations to large \emph{foundation world models} that learn physics and forecast future states from extensive datasets to support planning. AICON departs fundamentally from these predictive approaches. Rather than simulating future trajectories to choose actions, AICON encodes environmental regularities—such as relationships between sensory inputs and physical states—within a dynamic network of recursive estimators and active interconnections. Instead of explicitly predicting \( s_{t+1} \) for planning, AICON exploits this structure to compute gradients that directly steer the system toward immediate subgoals, replacing costly look-ahead search with efficient gradient descent. As a result, complex sequential behaviors emerge from continuous adaptation to real-time feedback. Because AICON does not roll out its model over long horizons, it is not exposed to compounding prediction errors in the way long-horizon predictive models are; its errors are those of its current estimates, which sensing continuously corrects.

	%----------------------------------------------------------------------
	\subsubsection{Causal Graphs}
	%----------------------------------------------------------------------

	A causal graph represents causation between variables as a directed acyclic graph. Nodes encode variables and edges encode the causal relationship between them. This structure enables reasoning about the effect of actions (interventions) or different outcomes of hypothetical scenarios (counterfactuals). This is especially useful for planning and model-based control where we want to model the consequences of possible actions.
	Like causal graphs, AICON encodes world regularities rather than task-specific rules or behavior. However, AICON can encode a greater variety of regularities and leverages them directly for action generation.
	First, AICON encodes regularities as undirected constraints between states, because information can flow in both directions. Therefore, AICON regularities include causal relationships and more general relationships such as geometric constraints.
	Second, AICON uses these regularities to generate behavior by propagating gradients through the network. Therefore, AICON is entirely reactive and does not require simulating the outcomes of different actions.
	Causal models answer explicit questions about interventions or counterfactuals, while AICON generates interventions that achieve a desired goal.

	%======================================================================
	\section{Biological Plausibility \& Modeling}\label{sec:biology}
	%======================================================================
	AICON was designed as a framework for robot behavior, but its core principles turn out to echo how biological systems are organized and how biological agents behave. This is not a coincidence: both robots and organisms must solve the same fundamental problem of acting robustly in a complex, partially observable world. The parallels run from architectural principles down to specific behavioral signatures, and have motivated us to use AICON as a modeling tool in its own right. We organize these parallels into three parts: structural similarities, behavioral similarities, and applications of AICON as a model of biological behavior.

	%----------------------------------------------------------------------
	\subsection{Biological Similarities in Structure}
	%----------------------------------------------------------------------
	Several of AICON's core architectural principles have direct biological counterparts. The most fundamental is the use of recursive estimation: biological perception is not a feedforward snapshot but a continuous, temporally integrated process in which prior beliefs are updated by incoming evidence---exactly the structure of a Bayesian filter~\cite{knill2004bayesian,rao1999predictive}. Related to this, AICON's distributed, hierarchical organization, in which components at different levels of abstraction exchange information bidirectionally, mirrors well-documented properties of the primate visual system, where signals flow not only bottom-up from sensors to higher areas but also laterally and top-down~\cite{mumford1992computational,felleman1991distributed,martinmartin2022coupled}. The reverse hierarchy theory of biological perception~\cite{ahissar2004reverse} describes exactly this pattern, and experimental evidence from fMRI and psychophysics supports the idea that top-down feedback serves as a prior to interpret ambiguous sensory signals~\cite{kersten2004object}.

	AICON's active interconnections---components that adjust the information they pass based on the current state of the system---find a biological analogue in the observation that robustness across biological systems seems to arise from context-sensitive information exchange rather than fixed wiring~\cite{li2024biologically}. This principle appears at multiple scales: in alternative splicing, where identical genetic building blocks produce different protein structures depending on their context; in gap junctions between cells, which support diverse and adaptive intercellular communication; and in cross-organ interactions in transplantation medicine~\cite{li2024biologically}. The common thread is that components are set up to interact in flexible, state-dependent ways, and that this flexibility is a precondition for robustness.

	The parallels extend to action generation as well: human motor systems are known to stabilize task-relevant dimensions while leaving null dimensions free for secondary objectives~\cite{scholz_uncontrolled_1999,latash_motor_2002}, a principle that maps directly onto AICON's nullspace formulation~\cite{mengers2026riding}. There is also neurophysiological evidence for projection-based control in motor cortex and prefrontal areas~\cite{churchland_neural_2012,mante_context-dependent_2013}, and behavioral evidence that humans select among multiple competing motor strategies in a way that resembles steepest gradient selection~\cite{akulin_sloppy_2019,mengers2025noplan}.

	%----------------------------------------------------------------------
	\subsection{Biological Similarities in Behavior}
	%----------------------------------------------------------------------
	Beyond structure, the behavior that emerges from AICON mirrors human behavior without being originally designed with that intention. Humans solving sequential manipulation tasks tend to rely on real-time feedback rather than explicit planning, taking epistemic actions~\cite{kirsh_distinguishing_1994} to gather information unless task constraints force them to plan ahead~\cite{waldron_influence_2011}. AICON does the same: it acts on the steepest available gradient given the current world state, which naturally produces information-gathering behavior when uncertainty is high~\cite{mengers2025noplan,bohg_interactive_2017}. The suboptimalities this introduces---limited look-ahead, sensitivity to goal hierarchies---match documented patterns in human performance on tower-based tasks~\cite{kaller_reviewing_2011,mckinlay_planning_2008}.

	We have verified these parallels more directly in a study of the Tower of London task comparing AICON's behavior to that of healthy participants and Parkinson's patients~\cite{migacev2026planhumanreactiverobotics}. Parkinson's patients are known to struggle specifically with goal hierarchies rather than with the raw number of moves required~\cite{mckinlay_planning_2008}. AICON exhibits the same asymmetry, for the same structural reason: both lack the machinery for deep goal-directed look-ahead while remaining capable of local, gradient-driven action. Interestingly, similar limitations in sequential foresight have been documented in great apes~\cite{tecwyn_novel_2013}, suggesting that AICON's behavioral profile may reflect something general about agents that act without explicit planning rather than a quirk of any particular implementation.

	%----------------------------------------------------------------------
	\subsection{Modeling Biological Behavior}
	%----------------------------------------------------------------------
	The structural and behavioral parallels have motivated us to use AICON not just as a robot controller but as a computational model of biological behavior. In vision science, AICON has been applied to model human perceptual illusions, reproducing effects that arise in human observers from the same principle of weakly decomposable inference that underlies AICON's architecture~\cite{battaje2024information}. The framework has also been used to model saccadic eye movements~\cite{mengers2025scanpath}, where AICON's distributed state estimation provides the perceptual substrate driving where and when the eyes move.

	At a larger scale, AICON has been applied to collective behavior, i.e., groups of agents coordinating under limited communication bandwidth, noise, and intermittency~\cite{mengers2024leveraging,mengers2026mechanistic}. Because AICON explicitly represents uncertainty, it handles degraded communication gracefully, and the resulting collective dynamics show qualitative similarities to naturally occurring collective motion such as milling in fish schools. The Tower of London study, the perceptual modeling work, and the collective behavior results together suggest that AICON's principles may generalize from individual perception and action to emergent group behavior---a direction we find genuinely exciting and one we are actively pursuing.

	%======================================================================
	\section{Limitations}\label{sec:limitations}
	%======================================================================

	AICON is a powerful framework for a specific class of problems, but understanding where it works well requires being equally clear about where it does not. Most of the limitations below are not implementation details to be engineered away---they reflect the fundamental commitments AICON makes in exchange for its robustness and reactivity.

	%----------------------------------------------------------------------
	\subsection{No Optimality Guarantees}
	%----------------------------------------------------------------------

	AICON selects the locally steepest gradient at each moment rather than searching over future action sequences. This makes it fast and reactive, but it means that AICON finds \emph{a} solution rather than the \emph{best} solution. A planner or MDP solver operating under correct models can, in principle, find time-optimal trajectories or minimize cumulative cost over a horizon. AICON cannot. In practice, the solutions AICON finds are often good enough---the real world rarely rewards the theoretically optimal plan because model errors compound over long horizons anyway---but for applications where optimality matters and the environment is well-specified, search-based methods remain the right tool.

	%----------------------------------------------------------------------
	\subsection{A Structural Ceiling on Solvable Problems}\label{sec:lim_ceiling}
	%----------------------------------------------------------------------

	AICON resolves subgoals by following the steepest gradient given the current world state, and resolves conflicts between simultaneously active goals using nullspace projections~(\Cref{sec:nullspace}). Both mechanisms operate on the current state: they ask what is best to do \emph{now}, not what sequence of states leads to a good outcome later.

	This creates a principled boundary on the class of solvable tasks~\cite{mengers2025noplan,mengers2026riding}. Conflicts between goals that are both active in the current state can be resolved reactively. Conflicts that only emerge through a future sequence of states---where a correct action now looks wrong given only the present situation---cannot. The Tower of Hanoi is the canonical example: moving a disk from its current position requires first moving another disk out of the way, but the gradient of the final goal does not reveal that this preparatory step is needed. Similarly, tasks involving irreversible actions with delayed consequences, or those requiring a deliberate temporary setback, fall outside AICON's reach without external planning assistance~\cite{mengers2025noplan}.

	This is not a flaw to be fixed but a reflection of the large-world/small-world tradeoff at AICON's core: by committing to continuous sensing rather than long-horizon anticipation, AICON gains robustness to model errors at the cost of look-ahead.

	%----------------------------------------------------------------------
	\subsection{Chattering Between Gradient Paths}\label{sec:lim_chattering}
	%----------------------------------------------------------------------

	The steepest-path rule of \Cref{eq:aicon_steepest} is a hard selection. When two paths have nearly equal gradient magnitudes but point in different directions in action space, small changes in the beliefs, including measurement noise, can flip the selection back and forth from one time step to the next. The resulting behavior chatters between two subgoals and may make little progress on either. The priority ordering of the nullspace composition (\Cref{sec:nullspace}) follows gradient magnitudes as well and is susceptible in the same way. In practice, standard measures keep this under control: \citet{mengers2026riding} accept a reordering only once a competing gradient exceeds the current one by a margin, low-pass filter the resulting control signal, and use different thresholds for entering and leaving nullspace exploration, so that the system does not immediately fall back into it. These parameters shape how sharply transitions occur rather than what the system does; across sweeps of margin, softmax temperature, and filter constant, success on the pushT task, in which the robot pushes a T-shaped block into a target pose, stayed between 94\% and 100\%~\cite{mengers2026riding}. Still, they are additions with their own parameters rather than consequences of the framework itself, and in fully deterministic settings residual limit cycles can remain, which small amounts of noise break~\cite{mengers2026riding}.

	%----------------------------------------------------------------------
	\subsection{Overconfidence from Dropped Cross-Correlations}\label{sec:lim_overconfidence}
	%----------------------------------------------------------------------

	AICON keeps a separate belief per state and does not track the cross-correlations between them (\Cref{sec:formalisms}). In a tree-structured graph, this approximation is benign. In a graph with loops, however, information from one source can reach an estimator along two different paths, and the estimator will treat the two arrivals as independent evidence. This double counting makes beliefs overconfident: uncertainties shrink faster than the evidence warrants. Because AICON's gains, and through them the gradient paths (\Cref{eq:readout}), depend on these uncertainties, overconfidence affects both perception and behavior. Conservative fusion rules such as Covariance Intersection (\Cref{tab:aicon_table}) avoid double counting at the cost of discarding some genuine information; choosing where in a graph such rules are needed is currently left to the designer.

	%----------------------------------------------------------------------
	\subsection{Generalization is Bounded by What is Encoded}\label{sec:lim_generalization}
	%----------------------------------------------------------------------

	AICON generalizes well within the regularities encoded in its graph---it recomposes existing estimators and interconnections to handle novel configurations, sensor readings, or task variants. But this generalization has a ceiling~\cite{battaje2026building}: if a situation depends on a regularity not represented anywhere in the graph, no recomposition of existing components can substitute for it. A robot that has never been given a model of friction cannot reason about tasks where friction is decisive, no matter how well its other components are specified.

	This is a meaningful distinction from learned models, which may acquire implicit representations of regularities from data even when those regularities were not explicitly designed in. AICON's generalization is more legible---you can inspect exactly what it knows---but also more bounded. Extending it to a new domain always requires asking what regularities that domain depends on and whether they are represented.

	%----------------------------------------------------------------------
	\subsection{The Graph Must Be Built, and Built Correctly}
	%----------------------------------------------------------------------

	AICON removes the need to hand-specify controllers, transitions, or action sequences. It does not remove the need for careful modeling. Every recursive estimator must have an appropriate belief representation, forward model, and where possible a measurement model. Every relevant interdependency between quantities must be encoded as an active interconnection. A missing interconnection does not raise an error---it silently removes an information channel or a gradient path, and the resulting behavior degrades in ways that can be surprisingly hard to trace. A misspecified constraint that is too tight over-constrains the system and causes erratic behavior; one that is too loose allows the estimators to drift and the gradients to weaken.

	In practice, building a correct AICON graph requires domain knowledge and iteration~(\Cref{sec:howto}). The good news is that failures are typically interpretable---missing gradient paths manifest as the system getting stuck at a specific world state, and missing information channels manifest as estimation uncertainty that doesn't resolve as expected---and fixes are local, since AICON's modularity means that adding a new component does not require redesigning the rest of the graph. But the modeling burden is real and should not be understated.

	%----------------------------------------------------------------------
	\subsection{Scope: Agents Interacting with the World}
	%----------------------------------------------------------------------

	Finally, AICON is a model of agents interacting with the world at a behavioral level of abstraction---above physics and chemistry, in the domain of poses, contacts, grasp states, and task-relevant relational properties. It is not a model of low-level signal processing, molecular dynamics, or phenomena where the relevant quantities are not interpretable as states of an interacting agent. The biological parallels we draw in \Cref{sec:biology} are meaningful precisely because biological systems are also agents interacting with the world; they would not transfer to domains where this framing does not apply.

	%======================================================================
	\section*{Acknowledgements}
	%======================================================================
	
	We thank David Kehrer, Alexander Koenig, Ole Liebich, Michael Migacev, and James Reynaldo for working through early versions of this tutorial and for their valuable feedback.
	
	We gratefully acknowledge funding by the Deutsche Forschungsgemeinschaft (DFG, German Research Foundation) under Germany's Excellence Strategy -- EXC 2002/2 ``Science of Intelligence'' -- project number 390523135. This work has been partially supported by the German Federal Ministry of Research, Technology and Space (BMFTR) under the Robotics Institute Germany (RIG).
	
	Claude Opus 5.5 (Anthropic) was used to help edit the TikZ code for \Cref{fig:intro:aicon_vs_others} and to improve the language and coherence of the text. The authors reviewed all suggestions and take full responsibility for the content.
	
	%======================================================================
	\section*{Author Contributions}\label{sec:contributions}
	%======================================================================
	
	\begingroup
	\renewcommand{\arraystretch}{1.4}
	\begin{tabularx}{\textwidth}{@{}l X@{}}
		\toprule
		\textbf{Author} & \textbf{Contributions} \\
		\midrule
		Vito Mengers & Wrote the first draft of the entire document and, in the current draft, contributed to the sections on the framework, the guide to building AICON graphs, the biology section, and the limitations; edited the entire draft. \\
		Adrian Pfisterer & Supported the first draft; contributed substantially to the sections on the AICON paradigm and framework, the interfaces with learning, the overview of applications, and the comparison to other approaches. \\
		Patrick Lowin & Wrote parts of the introductory example, the reinforcement learning interface, the perception overview, and the comparison to other approaches; edited the document for coherence and flow. \\
		Pu Xu & Wrote the first and second drafts of the introductory example contrasting planning, control, and AICON; edited the comparison to other approaches. \\
		Oliver Brock & Wrote the initial introduction and provided extensive high-level feedback throughout. \\
		\bottomrule
	\end{tabularx}
	\endgroup

	\bibliographystyle{IEEEtranN}
	\bibliography{references}

\end{document}